%% file: DynamicISP_iccv_arxiv_ver7.tex
\documentclass[10pt,letterpaper,twocolumn]{article}
\usepackage[latin9]{inputenc}
\usepackage{array}
\usepackage{multirow}
\usepackage{tipa}
\usepackage{amsmath}
\usepackage{amssymb}
\usepackage{graphicx}
\usepackage[unicode=true,
 bookmarks=false,
 breaklinks=false,pdfborder={0 0 1},backref=section,colorlinks=false]
 {hyperref}
\hypersetup{
 pagebackref,breaklinks,colorlinks}

\makeatletter

\pdfpageheight\paperheight
\pdfpagewidth\paperwidth

\DeclareTextSymbolDefault{\textquotedbl}{T1}
\providecommand{\tabularnewline}{\\}

\usepackage{iccv}
\usepackage{times}
\usepackage{epsfig}
\usepackage{graphicx}

\usepackage{import}

\iccvfinalcopy 

\def\iccvPaperID{11628} 

\ificcvfinal\fi

\makeatother

\begin{document}

\title{DynamicISP: Dynamically Controlled Image Signal Processor \\for Image
Recognition}

\author{
Masakazu Yoshimura\quad Junji Otsuka\quad Atsushi Irie\quad Takeshi Ohashi\\
Sony Group Corporation\\
{\tt\small \{masakazu.yoshimura, junji.otsuka, atsushi.irie, takeshi.a.ohashi\}@sony.com}%
}

\maketitle
\begin{abstract}
Image Signal Processors (ISPs) play important roles in image recognition
tasks as well as in the perceptual quality of captured images. In
most cases, experts make a lot of effort to manually tune many parameters
of ISPs, but the parameters are sub-optimal. In the literature, two
types of techniques have been actively studied: a machine learning-based
parameter tuning technique and a DNN-based ISP technique. The former
is lightweight but lacks expressive power. The latter has expressive
power, but the computational cost is too heavy on edge devices. To
solve these problems, we propose \textquotedblleft DynamicISP,\textquotedblright{}
which consists of multiple classical ISP functions and dynamically
controls the parameters of each frame according to the recognition
result of the previous frame. We show our method successfully controls
the parameters of multiple ISP functions and achieves state-of-the-art
accuracy with low computational cost in single and multi-category
object detection tasks.
\end{abstract}

\section{Introduction}

\label{sec:intro}

The image signal processor (ISP) is an important component of modern
digital cameras. It converts raw outputs of the sensors, RAW images,
into commonly used standard RGB (sRGB) images, and is composed of
many functions because of its multiple roles. For example, it calibrates
the color sensitivity of each sensor with a color correction matrix,
and improves image quality with demosaicing, denoising, and sharpening.
Auto-exposure, auto-white balance, and tone mapping are necessary
to mimic the adaptive and responsive characteristics of the human
eye to produce images that closely resemble the appearance perceived
by humans. The human eye cancels out the environment's luminance with
dark or light adaptation and also cancels out the environment's color
with color adaptation. As to the response characteristics, it is said
that the human eye non-linearly responds to light intensity roughly
obeying $y=x^{1/3}$ \cite{stevens1957psychophysical}. These functions
for human eye adaptation have another role. Optimizing these functions
for compression enables conversion from high dynamic range signals
in the real world to 8-bit sRGB images with little information loss.
The real world has the range from 0.0001 $[cd/m^{2}]$ (starlight)
to 1.6 billion $[cd/m^{2}]$ \cite{reinhard2010high} (direct sunlight).
We don't go any further, but ISPs have more functions, like lens shading
correction, dehazer, and bad pixel correction.

ISPs are also needed for image recognition tasks. DNN-based image
recognition, which has achieved excellent results recently, shows
even better performance with ISPs \cite{hansen2021isp4ml,buckler2017reconfiguring}.
However, ISP hyperparameters commonly require manual tuning by experts,
which consumes considerable time \cite{mosleh2020hardware}. The tuning
is necessary for each sensor. It is ineffective and might be sub-optimal,
especially for image recognition tasks. Easy-to-see images for humans
are not equal to easy-to-recognize images. In fact, recent works demonstrate
that the hyperparameter tuning for image recognition improves the
accuracy \cite{mosleh2020hardware,yu2021reconfigisp,robidoux2021end,tseng2019hyperparameter}.

\begin{figure}
\centering

\def\svgwidth{1.0\columnwidth}

\scriptsize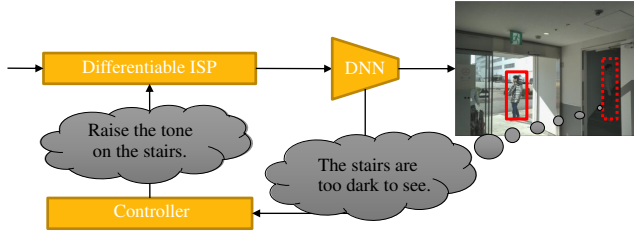

\caption{The concept of proposed DynamicISP. The parameters of the ISP are
controlled dynamically based on what the downstream recognition model
felt.}

\label{fig:concept}
\end{figure}

To achieve a better ISP for image recognition, we believe it is important
to improve the expressive power of the ISP since most ISP functions
are classical static functions. One solution is a DNN encoder-decoder-based
ISP; many works indeed achieved state-of-the-art accuracy in image
recognition performance \cite{diamond2021dirty,morawski2022genisp}
and perceptual quality \cite{punnappurath2022day,liu2022deep}. However,
the computation costs are enormous, especially when high-resolution
outputs were required. Devices such as smartphones have limited computational
resources and require an efficient ISP.

Based on the above, we propose ``DynamicISP,'' which consists of
multiple conventional ISP functions, but their parameters are controlled
dynamically per image. Although they are classical functions, their
dynamic control compensates for their expressive power. Our controller
is based on what the downstream recognition model felt and desired
at the previous time step as shown in Fig. \ref{fig:concept}. Specifically,
our method predicts appropriate ISP parameters for the following frame
based on intermediate features of the downstream recognition model. 

In terms of dynamic control, dynamic neural networks have been actively
studied \cite{yang2019condconv,chen2020dynamicc} and a similar method,
named NeuralAE, that controls auto-exposure with DNNs has recently
been published \cite{onzon2021neural}. Our method successfully controls
the parameters of multiple ISP functions, which is difficult even
with these successful dynamic control methods. In addition, despite
its low computational cost, our method achieves state-of-the-art accuracy
compared to encoder-decoder based DNN ISPs.

Our contributions are as follows:
\begin{itemize}
\item A novel ISP for image recognition that utilizes classical functions
but dynamically controls their parameters based on what the downstream
image recognition model senses.
\item A residual output format of parameters, which decomposes the control
into static tuning to average good parameters for the entire data
set and further pursuit of the best parameters per image. It eases
the difficulty of controlling complex functions.
\item A latent update style ISP controller that manages multiple ISP functions.
It generates shared latent variables which indicate how to change
the RAW images with the entire ISP pipeline; then, after applying
each ISP function, sequentially updates them to contain information
on what the remaining functions should do considering upstream functions.
\item End-to-end optimization method from ISP to image recognition model.
A proposed \emph{parameter initializer} module enabled the control
of complex functions. 
\item Diverse evaluations show the effectiveness of the proposed DynamicISP.
It achieves state-of-the-art accuracy despite the low computational
cost.
\end{itemize}

\section{Related Works}

\subsection{Image Signal Processor}

In recent years, many ISPs that utilize machine learning have been
proposed. They can be broadly classified into two categories: methods
that tune the classical ISP functions and methods that replace them
with a DNN-based encoder-decoder model.

In the first category, they tune the existing ISP parameters using
machine learning techniques. It defines perceptual loss or image recognition
loss, depending on the goal, to increase perceptual quality or recognition
accuracy. In some cases, evolutionary algorithms \cite{pavithra2021automatic,mosleh2020hardware,hevia2020optimization}
or other search methods \cite{nishimura2018automatic} are applied
because most existing ISPs are black-box and the search space of the
parameters have multiple local minima \cite{tseng2019hyperparameter}.
However, these methods optimize only ISP parameters for a fixed downstream
part, and the performance gain is limited. To partly address the problem,
a work proposes a joint training schedule wherein ISP and downstream
CNN are optimized alternately while the other is fixed \cite{robidoux2021end}.
Other methods optimize the parameters with gradient-based backpropagation
\cite{wu2019visionisp,onzon2021neural,yoshimura2022RAWAug}. Although
we mentioned NeuralAE \cite{onzon2021neural} controls auto-exposure
with DNN, ISP parameters are trained with backpropagation as static
parameters. ReconfigISP \cite{yu2021reconfigisp} not only optimizes
the hyperparameters but also explores functions by a neural architecture
search method \cite{liu2018darts}. It is important to note that the
computational costs of this category are low because their ISPs for
inference time consist of classical functions. We argue that the problem
with these methods is the lack of expressive power compared with the
other category.

In the other category, methods creates an ISP with encoder-decoder-based
DNN. Recently, tremendous model architectures are proposed as an ISP
function replacement, such as denoiser \cite{zhang2022idr,monakhova2022dancing},
color constancy \cite{ono2022degree,afifi2021cross}, and deblurring
\cite{zhang2022pixel,whang2022deblurring}. Some works propose generic
models that can replace various functions with the same architecture
\cite{tu2022maxim,chen2021hinet}, while several others replace the
whole ISP pipeline with DNN \cite{punnappurath2022day,morawski2022genisp,diamond2021dirty,liu2022deep}.
Though DNN-based ISPs and ISP functions achieve state-of-the-art,
the computational cost remains an issue especially when high-resolution
outputs are necessary. Currently, smartphones are one of the major
use cases of ISPs, thus low computational cost ISPs are highly desired.

Based on the above, we propose a dynamic ISP control method that doesn't
match either of these two categories. By dynamically controlling parameters,
classical ISPs are empowered with a minimal computational cost. Our
method is possible to optimize from ISP to image recognition end-to-end,
going beyond the usual individal optimization of the ISP and the recognition
model.

\subsection{Low-Light Image Enhancement}

\begin{figure*}
\centering

\def\svgwidth{1.9\columnwidth}

\scriptsize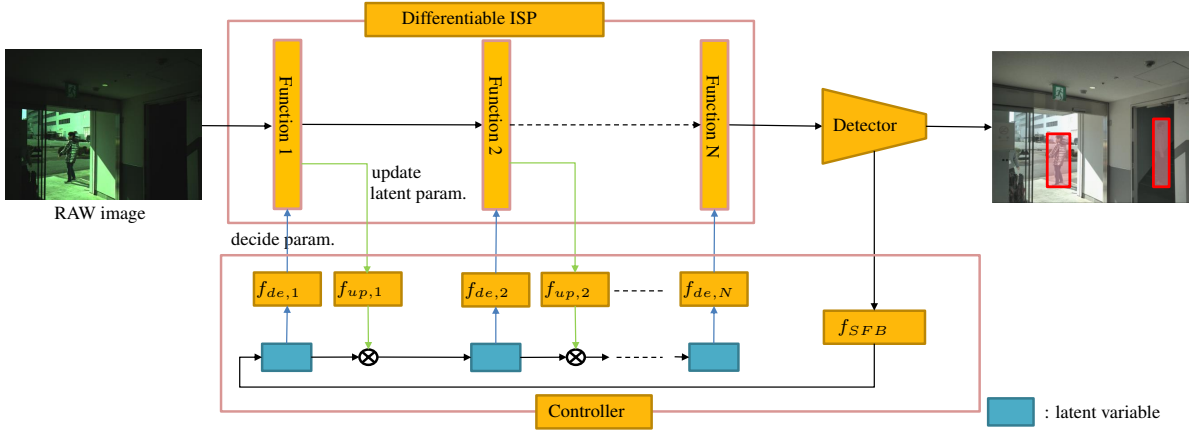

\caption{The proposed latent update style controller to control multi-layers.}

\label{fig:latent_update}
\end{figure*}

It is said that conventional ISPs tend to be poor in a low-light environment
\cite{guo2020zero,jiang2021enlightengan,zhang2019kindling}. Many
works focus on improving perceptual quality or recognition accuracy
in low-light environments. Most of them try to enhance the brightness
of pre-captured sRGB images \cite{lv2018mbllen,zhang2019kindling,jiang2021enlightengan,wei2018deep,guo2020zero,ma2022toward}.
These are very useful when improving already saved sRGB images but
some information is already vanished due to the lossy conversion from
RAW to sRGB. Other works optimize ISP for dark environment \cite{punnappurath2022day,morawski2022genisp,liu2022deep,diamond2021dirty}.
Several works train recognition models to recognize RAW images, which
have the richest information \cite{Hong2021Crafting,schwartz2021isp}.

Our first objective is to establish an ISP that can be used in any
environment covering from low light to dazzling, high dynamic range
(HDR), and even blurry. However, our DynamicISP can be useful when
targeting only dark scenes because optimal ISP parameters can be different
per environment even if the luminance is the same. Hence, we also
evaluate our method on low-light image recognition.

\subsection{Dynamic Parameter Control}

The dynamic model is one of the current hot research topics. Attention-based
feedforward control is successful in dynamically controlling DNN parameters.
Several works control the amplification values of the convolution
weights or PReLU \cite{he2015delving} using a separate attention
module fed features just before the convolution or PReLU \cite{yang2019condconv,chen2020dynamicc,chen2020dynamic}.
These works prove the effectiveness of controlling partial DNN parameters.

However, these attention-based feedforward controls are difficult
to apply to the parameters of the preprocessing part - there are no
rich features before the target functions. NeuralAE \cite{onzon2021neural}
solves these situations by introducing a feedback circuit in which
an intermediate feature of the detector's backbone from the previous
time step is used to control the exposure.

Although our ISP control can be based on feedback control as NeruralAE
does, the control of multiple ISP functions with a controller has
never been realized to the best of our knowledge; it fails when we
simply apply the existing NeuralAE's method. Therefore, we propose
the feedback control method to control parameters of multiple ISP
functions.

\section{Methodology}

In this section, the proposed dynamic ISP control and its training
method are presented.

\subsection{Dynamic ISP Control}

Our problem setting for ISP control is to predict the next frame $(t+1)$
parameters $P=\{P_{l}|l=1,...,L\}$ of ISP functions $I(x)=\{I_{l}\left(x\right)|l=1,...,L\}$,
where $L$ is the number of ISP functions, to boost visibility for
machine vision. Each function $I_{l}\left(x\right)$ contains $N_{l}$
parameters $P_{l}=\{p_{l,n}^{t+1}|n=1,...,N_{l}\}$. In a static frame
scenario, we let the next frame $(t+1)$ just the same image as the
previous frame $t$. The parameters $P$ are estimated based on encoded
intermediate feature $X_{t}$ of the downstream recognition model
in a feedback control format according to NeuralAE \cite{onzon2021neural}.
Specifically, we use the ``\emph{Semantic Feature Branch}'' $f_{SFB}(X_{t})$
proposed in NeuralAE to encode the features. While NeuralAE adds another
histogram-based RAW image encoder independent from the recognition
model, we did not use it because it was the bottleneck of the speed
in our implementation, and no accuracy improvement was noticed, particularly
when a large downstream recognition model was used.

\subsubsection*{Residual Output Format of Parameters\label{subsec:Residual-Output-Format}}

NeuralAE predicts the desired exposure by
\begin{equation}
p_{0,0}^{t+1}=f_{act}\left(f_{full}\left(f_{SFB}\left(X_{t}\right)\right)\right),\label{eq:1}
\end{equation}
 where $f_{full}$ and $f_{act}$ are fully connected layers and an
output activation layer respectively. Since ISP functions are not
as simple as the exposure function, which just multiplies the input,
we need to propose an elaborate output format to control them properly.
We introduce learnable parameter $\hat{p_{l,n}}$ as 
\begin{equation}
p_{l,n}^{t+1}=f_{act}\left(\hat{p_{l,n}}+f_{full}\left(f_{SFB}\left(X_{t}\right)\right)\right).\label{eq:2}
\end{equation}
 By this formulation, the $\hat{p_{l,n}}$ can be trained as an average
good value for the entire data set, while the controller only predicts
the gap between the average good value and the best value per image.
It eases the difficulty of controlling complex functions. In detail,
we define $f_{act}$ as 
\begin{equation}
f_{act}(x)=\left(p_{l,n,max}-p_{l,n,min}\right)\cdot sigmoid(x)+p_{l,n,min},\label{eq:4}
\end{equation}
 to search in $\left(p_{l,n,min},p_{l,n,max}\right)$. This formulation
is handy because we can easily set the search space of any hyperparameters
based on their domain. Some ISP parameters are limited in the range
they can take. In the experiments, wide search ranges were set to
allow free control within the ranges they can take.

\subsubsection*{Latent Update Style Controller\label{subsec:Latent-Update-Style}}

Both attention-based dynamic DNN \cite{yang2019condconv,chen2020dynamicc}
and NeuralAE \cite{onzon2021neural} predict parameters of only one
function by a controller. However, multi-layer ISP control is necessary
to predict the parameters of multi-layers. As modifying the parameters
in upstream functions affect the following functions, controlling
all functions is challenging. In fact, our experiments show that existing
methods are difficult to control multi-layers. We tackle this issue
with a proposed latent updated style controller. The proposal is illustrated
in Fig. \ref{fig:latent_update}. First, the main controller $f_{SFB}$
extracts what the downstream image recognition model felt and reacted
to the previous frame and converts the information into a latent variable
that contains what to do in overall functions. Then, after applying
each function, the latent variable is updated to contain information
on what to do in the remaining functions. The concept is realized
with the following formula: 
\begin{equation}
\begin{cases}
P_{l}=f_{de,l}\left(V_{l-1}\right)=f_{act,l}\left(\hat{P_{l}}+f_{full,l}(V_{l-1})\right)\\
V_{l}=f_{up,l}\left(P_{l},V_{l-1}\right)
\end{cases},\label{eq:5}
\end{equation}
 where $V_{l-1}$ is the latent variable for deciding the parameters
$P_{l}$ of function $I_{l}$ with $f_{de,l}$, and $f_{ud,l}$ updates
it according to what conversion is done in $I_{l}$. In our experiments,
we define $f_{ud,l}$ simply as 
\begin{equation}
f_{up,l}\left(P_{l},V_{l-1}\right)=f_{a}\left(P_{l}\right)\cdot V_{l-1},\label{eq:6}
\end{equation}
 where $f_{a}\left(P_{l}\right)$ consists of a normalization layer,
two fully connected layers, a ReLU activation between the fully connected
layers, then an activation, $y=5\cdot sigmoid(x)$. The normalization
layer is based on the range of the parameter search spaces: $\left(p_{l,n}^{t+1}-p_{l,n,min}\right)/\left(p_{l,n,max}-p_{l,n,min}\right)$
for each $n$. 

The latent update style disentangles the multi-layer control problem
setting without adding image encoders between the functions to convert
it into a single-layer control per each estimator problem. Adding
image encoders per function increases the computational cost compared
to our 1-D latent variable manipulations.

\subsection{Training Method\label{subsec:Training-Method}}

We mainly follow the two-frame training method of NeuralAE \cite{onzon2021neural}.
The first frame is input to the differentiably implemented ISP and
to a part of the image recognition backbone to get the intermediate
feature $X_{t}$ for ISP control. Second, the ISP is controlled for
the next frame. Third, the next frame is input to the ISP and the
image recognition model to compute a recognition loss. Finally, the
recognition loss for the second frame is backpropagated through all
paths above to train the ISP, controller, and recognition model end-to-end.

In NeuralAE, the exposure of the first frame is set to default, in
other words, the exposure of the first frame is set to the same value
used in data capturing. One major problem with ISP control is that
it consists of non-linear functions and drastically changes the original
RAW input depending on the parameters. To train the controller, it
is fundamental to properly manage the ISP parameters for the first
frame. To this end, we propose a \emph{parameter initializer} that
manages ISP parameters of the first frame at the training time.

\subsubsection*{Parameter Initializer\label{subsec:Parameter-Setter}}

\begin{figure}[h]
\centering

\def\svgwidth{1.0\columnwidth}

\scriptsize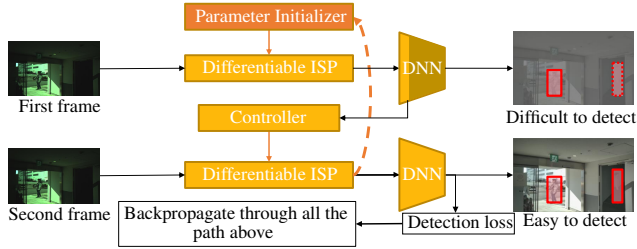

\caption{The proposed training method. The\emph{ parameter initializer} decides
the ISP parameters for the first frame based on what kind of parameters
were used for the second frames.}

\label{fig:parameter_setter}
\end{figure}

As mentioned above, this module sets ISP parameters of the first frame
at training time. The parameters are needed to be not always optimal
values for the input to train how to improve. However, it still should
be realistic values - learning parameter domain that will never be
used at inference time is inefficient and makes the training difficult.
Additionally, hand tuning the range for all parameters is difficult.
Therefore we propose a method to choose realistic parameters automatically
based on what kind of parameters was predicted as optimal for the
second frames in the previous iterations by the controller as illustrated
in Fig. \ref{fig:parameter_setter}. Our approach is to memorize the
estimated parameters in the most recent $M$ data in a buffer and
randomly use the value at the first frame. We also consider sampling
parameters from Gaussian distribution based on running mean and variance
of the past parameters. This approach assume the parameters follow
Gaussian distribution, but we are not sure it is true. Therefore,
we expect our approach can generate more plausible initial parameters
easily than other approaches which requires some assumptions or domain
knowledge.

\section{Evaluation}

\subsection{Datasets}

Although DynamicISP can be applied to any recognition task, it is
evaluated on detection tasks because of their wide usage. Two datasets
are used to check the robustness in different settings. One is a human
detection dataset \cite{yoshimura2022RAWAug}. This dataset aims to
detect humans in any environment while training only with simple environmental
images, as annotating difficult images is costly. Most of the test
images are taken in challenging environments: dark, HDR, and heavy
handshakes. They are taken with a RAW Bayer sensor. It has 18,880
easy images for the training set and 2,800 difficult images for the
test set.

The other is LODDataset \cite{Hong2021Crafting}, which contains 2230
14-bit low-light RAW images with eight categories of objects. This
dataset aims to detect multi-category objects in a low-light environment.

\subsection{ISP functions}

We implement five ISP functions, auto gain (AG), denoiser (DN), sharpener
(SN), gamma tone mapping (GM), and contrast stretcher (CS), in a differentiable
manner. The details are in the supplemental material. In the setting
of single-function ISPs, GM is used as an ISP since it is known to
improve image recognition accuracy the most \cite{hansen2021isp4ml}.
In the setting of multi-function ISPs, the combination of functions
is determined for each dataset after several experiments. It is known
that the optimal combination of functions varies depending on the
sensor, task, and environment \cite{yu2021reconfigisp}. In the future,
this process could be automated by combining our method with a neural
architecture search like ReconfigISP \cite{yu2021reconfigisp}.

\subsection{Ablation Studies on Human Detection}

\begin{figure*}
\centering

\setlength{\tabcolsep}{0.4pt} 


\begin{tabular}{cccccc}
\includegraphics[scale=0.142857]{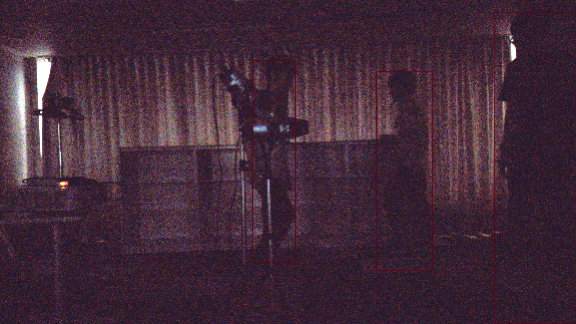} & \includegraphics[scale=0.142857]{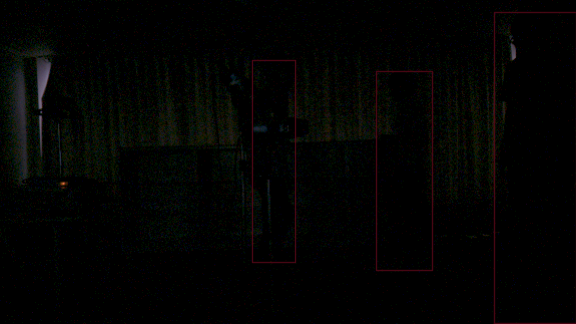} & \includegraphics[scale=0.142857]{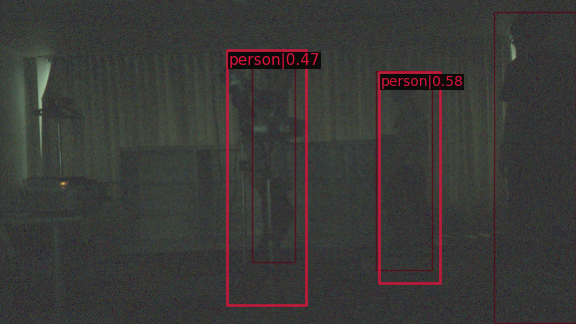} & \includegraphics[scale=0.142857]{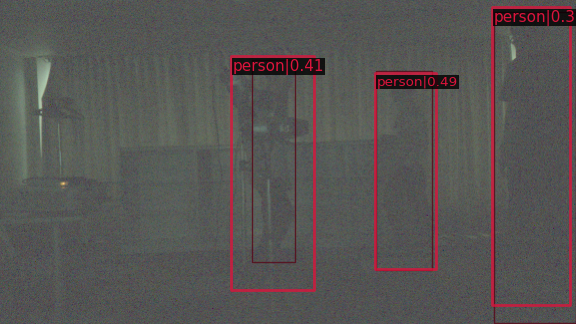} & \includegraphics[scale=0.142857]{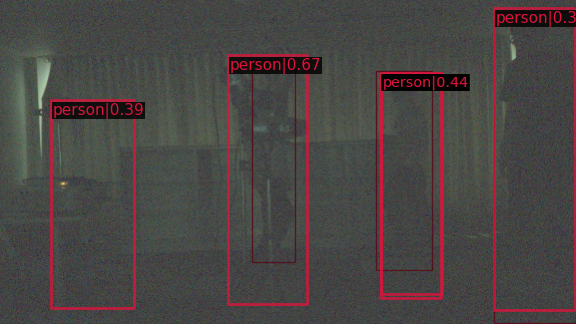} & \includegraphics[scale=0.142857]{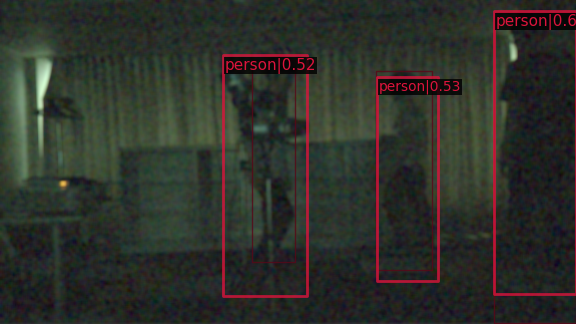}\tabularnewline
\includegraphics[scale=0.142857]{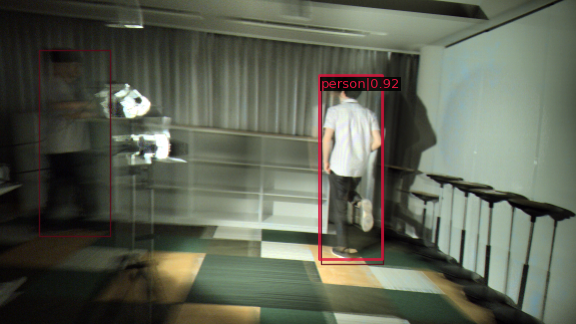} & \includegraphics[scale=0.142857]{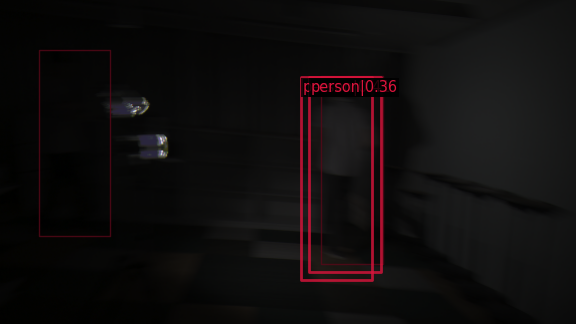} & \includegraphics[scale=0.142857]{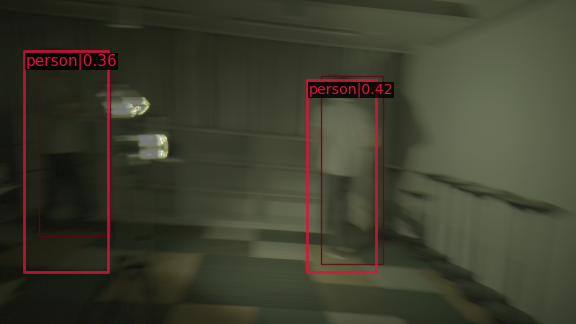} & \includegraphics[scale=0.142857]{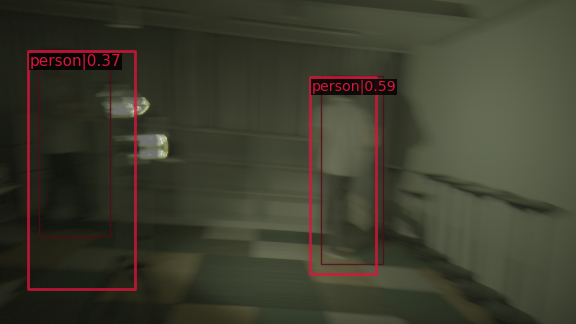} & \includegraphics[scale=0.142857]{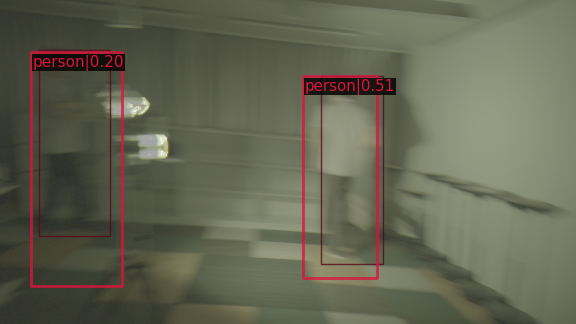} & \includegraphics[scale=0.142857]{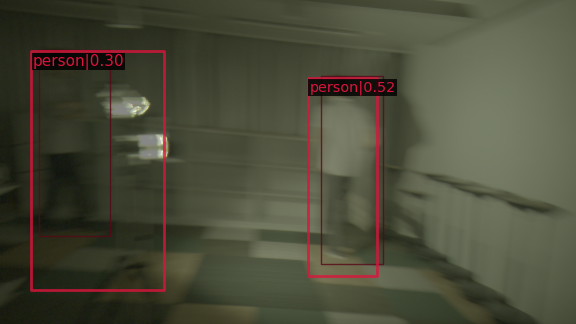}\tabularnewline
\includegraphics[scale=0.142857]{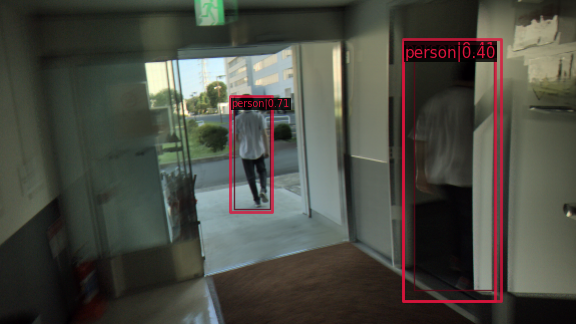} & \includegraphics[scale=0.142857]{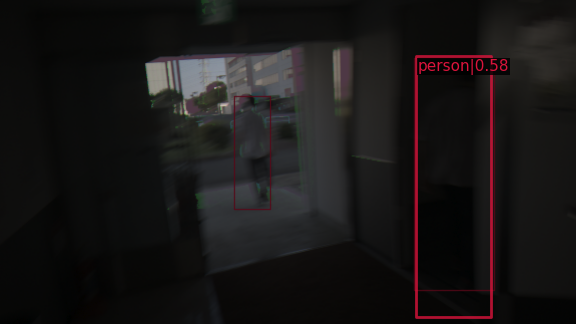} & \includegraphics[scale=0.142857]{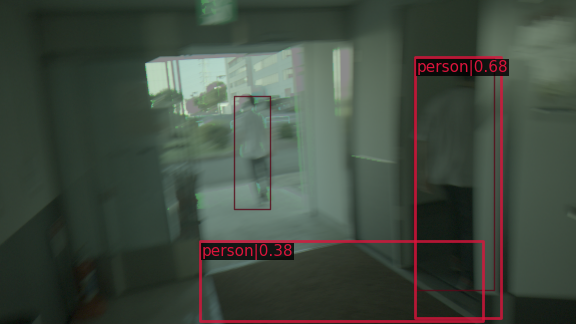} & \includegraphics[scale=0.142857]{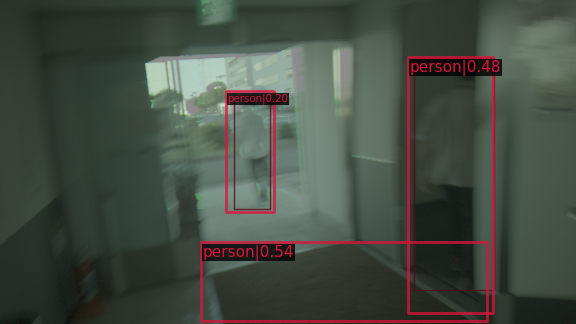} & \includegraphics[scale=0.142857]{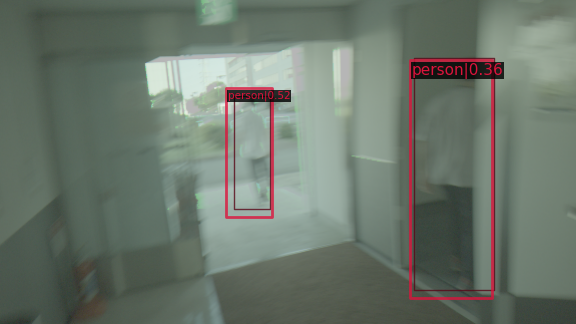} & \includegraphics[scale=0.142857]{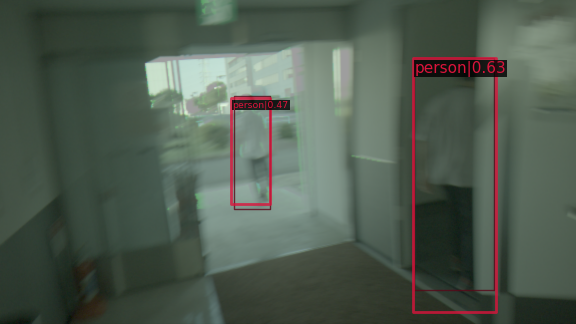}\tabularnewline
\includegraphics[scale=0.142857]{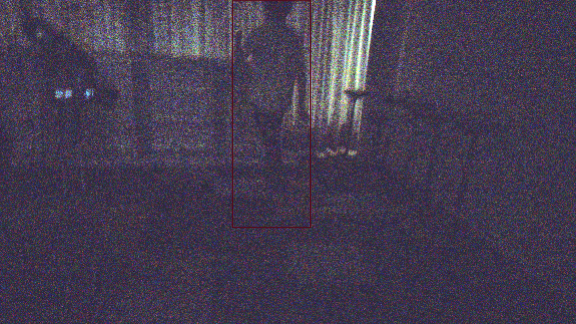} & \includegraphics[scale=0.142857]{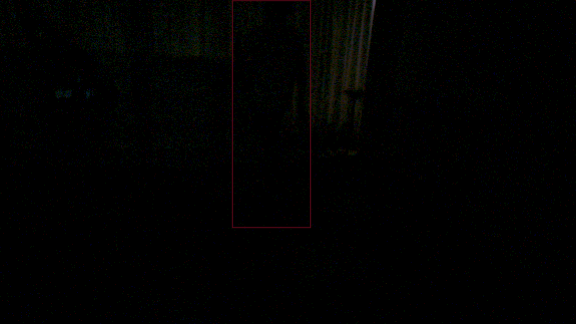} & \includegraphics[scale=0.142857]{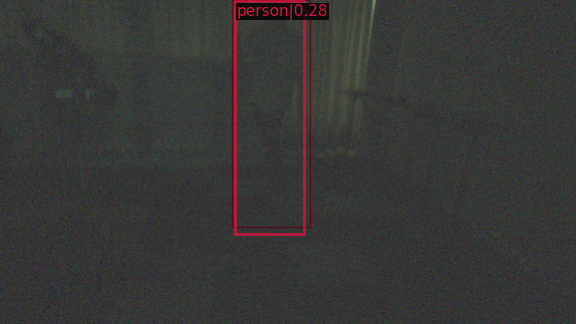} & \includegraphics[scale=0.142857]{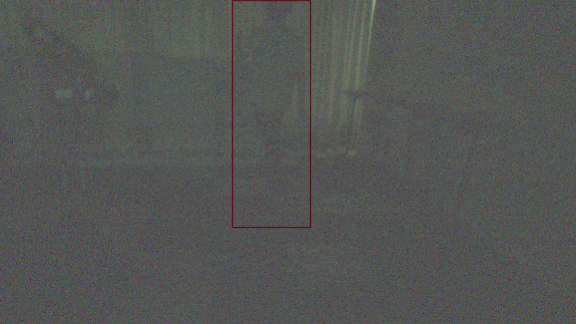} & \includegraphics[scale=0.142857]{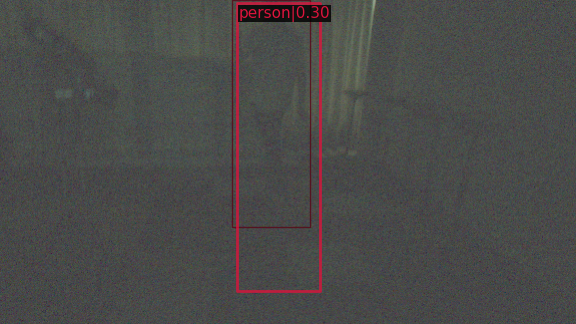} & \includegraphics[scale=0.142857]{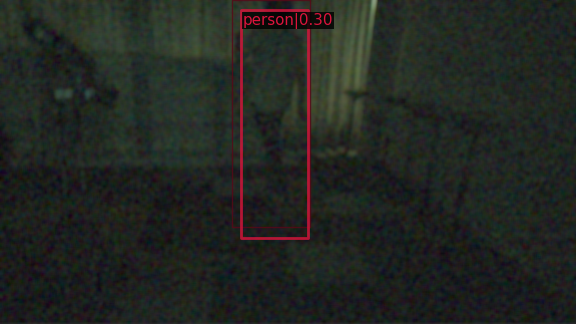}\tabularnewline
\scalebox{0.8}{%
\begin{tabular}{c}
sRGB\tabularnewline
(ISP=Black-box)\tabularnewline
\end{tabular}} & \scalebox{0.8}{%
\begin{tabular}{c}
DNN ISP \cite{punnappurath2022day} \tabularnewline
AWB+sGM+UNet(p)\tabularnewline
\end{tabular}} & \scalebox{0.8}{%
\begin{tabular}{c}
Rawgment \cite{yoshimura2022RAWAug}\tabularnewline
(ISP=GM)\tabularnewline
\end{tabular} } & \scalebox{0.8}{%
\begin{tabular}{c}
NeuralAE \cite{onzon2021neural}\tabularnewline
(ISP=GM)\tabularnewline
\end{tabular} } & \scalebox{0.8}{%
\begin{tabular}{c}
Ours\tabularnewline
(ISP=GM)\tabularnewline
\end{tabular}} & \scalebox{0.8}{%
\begin{tabular}{c}
Ours\tabularnewline
(ISP=DN+SN+GM+CS)\tabularnewline
\end{tabular}}\tabularnewline
\end{tabular}

\caption{The visualization result. We set an adequate confidence threshold
per model as the precision@0.5 becomes 80\% for a fair comparison.
The darker bounding boxes are ground truth, and the brighter bounding
boxes with captions are predictions.}

\label{fig:vizu}
\end{figure*}

Various ablation studies are performed using the human detection RAW
dataset \cite{yoshimura2022RAWAug}. We evaluate with the following
setting if not specified. TTFNet \cite{liu2020training} with ResNet18
\cite{he2016deep} backbone is used as the detector. The entire pipeline
is trained for 24 epochs with Adam optimizer \cite{kingma2014adam}
from randomly initialized weights using a cosine decay learning rate
schedule (0.001 magnitudes) whose maximum and minimum learning rates
are 1e-3 and 1e-6 with a linear warmup for the first 1,000 iterations
\cite{loshchilov2016sgdr}. The controller's learning rate is multiplied
by 0.1 from the base learning rate. After that, the model is additionally
trained for 24 epochs with the same learning rate scheduling, but
the ISP and controller are frozen the samely with NeuralAE \cite{onzon2021neural}.
This procedure slightly improves the accuracy from simple 48-epoch
training. The input size is set to $\left(576,\,352,\,3\right)$,
and Rawgment \cite{yoshimura2022RAWAug} is applied to bridge the
domain gap between the training data and challenging test data. During
the training phase, the identical frame is input twice instead of
using the continuous frame as described in Section \ref{subsec:Training-Method}
due to the rough frame rate of about 1 fps. In the evaluation phase,
on the other hand, we try both sequential input and identical input.
We set the channel size of the latent variable in the proposed controller
as 256. The accuracy is evaluated with the average AP@0.5:0.95 \cite{lin2014microsoft}
scores of four-time experiments with different random seeds.

\subsubsection*{Importance of Each Proposed Component}

\begin{table}[h]
\caption{Ablation studies about each proposed component. We add them one by
one to the controller of NeuralAE \cite{onzon2021neural}. The abbreviations
of the proposed method are as follows; RO: \emph{residual output format
of parameters} whose default parameters are static, RO+: \emph{residual
output format of parameters }whose default parameters are learnable,
PI: \emph{parameter initializer}, and LU: \emph{latent update style
controller}. The metrics are AP@0.5:0.95 {[}\%{]}.}

\begin{center}

\begin{tabular}{cccc|cc}
\hline 
 &  &  &  & \multicolumn{2}{c}{ISP components}\tabularnewline
\hline 
PI & RO & RO+ & LU & GM & DN+SN+GM+CS\tabularnewline
\hline 
 &  &  &  & 42.8 & 46.0\tabularnewline
$\checkmark$ &  &  &  & 45.8 & -\tabularnewline
$\checkmark$ & $\checkmark$ &  &  & 48.6 & -\tabularnewline
$\checkmark$ & $\checkmark$ & $\checkmark$ &  & \textbf{48.9} & 48.0\tabularnewline
$\checkmark$ & $\checkmark$ & $\checkmark$ & $\checkmark$ & - & \textbf{49.5}\tabularnewline
\hline 
\end{tabular}

\end{center}

\label{tab:val_components}
\end{table}

We add each proposed component one by one to the NeuralAE controller
\cite{onzon2021neural} to control two ISPs. One consisting of only
GM and the other consisting of DN, SN, GM, and CS. The reason we do
not use AG is that we tried to control AG, but for some reason, it
failed on this dataset. The result is shown in Table \ref{tab:val_components}. 

Even when controlling only the GM, the complexity of the GM function
makes it difficult to control with the NeuralAE method. The \emph{parameter
initializer} (PI) contributes significantly to alleviate the difficulty;
without PI, the detector's backbone has to deal with the pixel distributions
before and after ISP, degrading detection performance. A detailed
ablation study of the PI is shown in Table \ref{tab:param_setter}.
We compare three types of PIs. One is a uniform sampling from $\left(p_{l,n,min},\,p_{l,n,max}\right)$;
it learns how to control from all possible parameters. Another adds
Gaussian noise to the running mean of the previously used parameters
by tracking the running mean and running variance of used parameters
in the second frame. The last approach is our best method, which memorizes
the parameters for the 500 most recent data in a buffer and randomly
adopts the value in the first frame. It takes into account correlations
between parameters and does not approximate a Gaussian distribution,
thus covering more necessary and sufficient parameter regions. Our
buffer memory automatically fits the search space to the minimum required,
although it can still make the controller stable to the drastic environmental
change. The buffer memory creates a situation where the environment
is changed from day to night in a one-time step.

\begin{table}[h]
\caption{Ablation studies about the\emph{ parameter initializer}.}

\vspace{2mm}

\begin{center}

\scalebox{0.9}{

\begin{tabular}{c|c}
\hline 
 & AP@0.5:0.95\tabularnewline
\hline 
\hline 
None & 42.8\tabularnewline
uniform & 44.2\tabularnewline
running mean + Gaussian & 44.7\tabularnewline
buffer memory & \textbf{45.8}\tabularnewline
\hline 
\end{tabular}}

\end{center}

\label{tab:param_setter}
\end{table}

The \emph{residual output format of parameters} (RO) further eases
the difficulty of controlling complex functions by making the controller
output the difference from the averaging good parameters for the entire
dataset. For the normal RO, the static ISP parameter obtained by differentiable
tuning in Table \ref{tab:static_vs_dynamic} is used as a constant
$\hat{p_{l,n}}$. On the other hand, $\hat{p_{l,n}}$ in RO+ is jointly
optimized with the dynamic ISP control from scratch. The gain from
RO to RO+ is thought to be because the averaging good parameters for
the static ISP and DynamicISP is different. RO+ is practical as the
parameters are tuned automatically. 

For multi-layer control, the \emph{latent update style controller}
(LU) further improves the accuracy. All-at-one estimation of all the
parameters in the multi-layers is difficult and the proposed LU successfully
disentangles the problem setting.

\subsubsection*{Static Tuning V.S. Dynamic Control}

The proposed dynamic control is compared to static tuning approaches.
In this comparison, we use an ISP containing only a GM tone mapping
function because tone mapping is known to have the biggest impact
on machine vision \cite{hansen2021isp4ml}. One of the static tuning
approaches for comparison is a grid search \cite{yoshimura2022RAWAug}.
In this approach, the detector is trained per searched parameters.
Since it takes a huge time to train the detector per parameter set
of large search space, it searches for one parameter of the simplest
gamma function $y=x^{\frac{1}{\text{\textramshorns}}}$. The other
is a differentiable tuning approach to train the ISP and detector
end-to-end \cite{yoshimura2022RAWAug}. Table \ref{tab:static_vs_dynamic}
shows the effectiveness of our dynamic control. Although both static
ISPs are optimized for machine vision, they are just optimal for the
entire dataset and not optimal for each image. 

\begin{table}[h]
\caption{Static tuning V.S. dynamic control. The grid search method is the
value described in Rawgment performs\cite{yoshimura2022RAWAug} using
a simple gamma. The diff. tuning denotes differentiable tuning of
the static ISP parameters also described in Rawgment \cite{yoshimura2022RAWAug}.}

\begin{center}

\scalebox{0.9}{

\begin{tabular}{c|cc}
\hline 
 & \multicolumn{2}{c}{ISP components}\tabularnewline
\cline{2-3} \cline{3-3} 
 & GM & DN+SN+GM+CS\tabularnewline
\hline 
\hline 
grid search \cite{yoshimura2022RAWAug} & 45.3 & -\tabularnewline
diff. tuning \cite{yoshimura2022RAWAug} & 48.3 & 45.1\tabularnewline
dynamic control & \textbf{48.9} & \textbf{49.5}\tabularnewline
\hline 
\end{tabular}}

\end{center}

\vspace{-3mm}

\label{tab:static_vs_dynamic}
\end{table}

\subsubsection*{Sequential Control Availability}

The experiments above are conducted by inputting identical images
twice instead of sequential input. We now check the availability of
sequential inference without twice input. In this setting, optimal
parameters for the previous frame are used, resulting in possibly
suboptimal. The result is shown in Table \ref{tab:sequential_eval}.
Contrary to expectations, a slight improvement is observed, and it
might come from the fact that the twice input case is evaluated under
severe settings. The initial parameters are set as the moving average
of the parameters in the training phase and are far from the optimal
parameters per image. On the other hand, in the sequential settings,
the previous parameters can be already decent values since consecutive
frames are similar. From this experiment, efficient inference without
input twice is proved to be possible. Additionally, since the dataset
is captured at about 1 fps, and about half of it is taken with a shaken
camera, consecutive frames may have some disparities. Inference at
higher frame rates might further improve the performance of sequential
inference.

\begin{table}[h]
\caption{The availability of efficient sequential inference. The \textquotedblleft input
twice\textquotedblright{} is the same as all the experiments above.
It uses the same frame for determining ISP's parameters and detection.
On the other hand, in \textquotedbl input sequentially,\textquotedblright{}
optimal parameters for the previous frame is used for the next frame.}

\begin{center}

\scalebox{0.9}{

\begin{tabular}{c|cc}
\hline 
 & \multicolumn{2}{c}{ISP components}\tabularnewline
\cline{2-3} \cline{3-3} 
 & GM & DN+SN+GM+CS\tabularnewline
\hline 
\hline 
input twice & 48.9 & 49.5\tabularnewline
input sequentialy & \textbf{49.1} & \textbf{49.6}\tabularnewline
\hline 
\end{tabular}}

\end{center}

\vspace{-3mm}

\label{tab:sequential_eval}
\end{table}

\subsubsection*{Computational Efficiency}

\begin{table}[h]
\caption{The computational cost of the inference pipeline. $C$ is the computational
cost of the ISP. Any ISP configurations don't change the computational
cost of the controller very much; adding an ISP function increases
the cost by only about 4e-5 GFLOPS.}

\begin{center}

\scalebox{0.9}{

\begin{tabular}{c|ccc|c}
\hline 
 & \multicolumn{4}{c}{computational cost {[}GFLOPS{]}}\tabularnewline
\cline{2-5} \cline{3-5} \cline{4-5} \cline{5-5} 
 & ISP & controller & detector & overall\tabularnewline
\hline 
\hline 
w/o control & $C$ & - & 13.63 & 13.63+$C$\tabularnewline
input twice & 2$C$ & 0.02 & 15.99 & 16.02+2$C$\tabularnewline
input sequentialy & $C$ & 0.02 & 13.63 & 13.65+$C$\tabularnewline
\hline 
\end{tabular}}

\end{center}

\label{tab:computational_cost}
\end{table}

The computational efficiency is evaluated by profiling the FLOPS of
each component. As listed in Table \ref{tab:computational_cost},
the computational cost of our ISP controller is lightweight enough
to be ignored compared with the detector despite our choice of a lightweight
TTFNet detector with a ResNet18 backbone. Although our proposed LU
has a complicated computational graph as shown in Fig. \ref{fig:latent_update},
the computational cost does not change very much (an increase of about
4e-5 GFLOPS per each ISP function) thanks to the lightweight design
with 1-D latent variable manipulations. The computational cost of
the detector in the input twice setting is not doubled from other
settings because the whole part of the calculation for the first input
is not needed. If ISP functions are lightweight classical functions,
the dynamic ISP control can be efficient even as a single image detector
with the inputting twice setting.

\subsubsection*{Comparison with State-of-the-arts}

\begin{table*}
\caption{Comparison with state-of-the-arts. $C$ and $H$ are the computational
cost of the ISP and generating histograms of an image.}

\vspace{4mm}

\begin{center}

\scalebox{0.9}{

\begin{tabular}{cc|ccccc}
\hline 
\multicolumn{2}{c|}{ISP} & GFLOPS & \#params {[}M{]} & latency {[}ms{]} & w/o detector {[}ms{]} & AP {[}\%{]}\tabularnewline
\hline 
\hline 
w/o ISP &  & 13.63 & 14.20 & 7.0 & 0.0 & 32.8\tabularnewline
\hline 
\multirow{3}{*}{%
\begin{tabular}{c}
DNN\tabularnewline
based\tabularnewline
\end{tabular}} & UNet(s) & 21.96 & 21.96 & 13.8 & 6.8 & 28.7\tabularnewline
 & AWB+sGM+UNet(f) \cite{punnappurath2022day} & 21.96+C & 21.96 & 14.2 & 7.2 & 17.3\tabularnewline
 & AWB+sGM+UNet(p) \cite{punnappurath2022day} & 21.96+C & 21.96 & 14.2 & 7.2 & 42.0\tabularnewline
\hline 
\multirow{8}{*}{%
\begin{tabular}{c}
classical\tabularnewline
function\tabularnewline
based\tabularnewline
\end{tabular}} & sRGB (black box ISP) & 13.63 & 14.20 & - & - & 38.4\tabularnewline
 & diff. tuning (ISP=DN+SN+GM+CS) \cite{yoshimura2022RAWAug} & 13.63+C & 14.20 & 25.8$\spadesuit$ & 18.8$\spadesuit$ & 45.1\tabularnewline
 & NeuralAE (ISP=DN+SN+GM+CS) \cite{onzon2021neural} & 13.65+C & 15.66 & 26.0$\spadesuit$ & 19.0$\spadesuit$ & 45.2\tabularnewline
 & NeuralAE (ISP=GM)$\clubsuit$ \cite{onzon2021neural} & 13.65+C+H & 15.66 & 8.4 & 1.4 & 47.0\tabularnewline
 & Hardware-in-the-loop \cite{mosleh2020hardware} & 13.63+C & 14.20 & 8.1 & 1.1 & 47.1\tabularnewline
 & diff. tuning (ISP=GM) \cite{yoshimura2022RAWAug} & 13.63+C & 14.20 & 8.1 & 1.1 & 48.3\tabularnewline
 & NeuralAE (ISP=GM)$\dagger$ \cite{onzon2021neural} & 13.65+C+H & 15.66 & 18.3 & 11.3 & 48.5\tabularnewline
 & NeuralAE (ISP=GM) \cite{onzon2021neural} & 13.65+C & 15.66 & 8.4 & 1.4 & 48.6\tabularnewline
\hline 
\multirow{2}{*}{dynamic} & ours (ISP=GM) & 13.65+C & 15.66 & 8.3 & 1.3 & 49.1\tabularnewline
 & ours (ISP=DN+SN+GM+CS) & 13.65+C & 15.76 & 27.7$\spadesuit$ & 20.7$\spadesuit$ & \textbf{49.6}\tabularnewline
\hline 
\end{tabular}}

\end{center}

\scalebox{0.8}{

$\clubsuit$: The original augmentation in stead of Rawgment\cite{yoshimura2022RAWAug}
is used.

}

\scalebox{0.8}{

$\dagger$: Full version of NeuralAE including \emph{Global Feature
Branch}, which is a histgram-based encoder to support\emph{ Semantic
Feature Branch.}

}

\scalebox{0.8}{

$\spadesuit$: Our implementations of DN and SN have a lot of duplicate
calculations and are not optimized.

}

\label{tab:comp_sota}
\end{table*}

Various evaluations are performed against the state-of-the-art. The
latency is measured on an RTX2080 GPU. The result is shown in Table
\ref{tab:comp_sota}.

One category for comparison is DNN-based encoder-decoder ISPs. The
problem with the DNN ISP is that there are no input and ground truth
pairs to train them. As pointed out in \cite{punnappurath2022day},
creating clean ground truth for challenging noisy/blurry environment
data is difficult. So, three kinds of DNN ISPs are tried. One is UNet(s);
an UNet \cite{ronneberger2015u} based ISP , and the ISP and detector
are trained end-to-end from scratch. It means that the ISP is trained
only with the detection loss. The second is AW+sGM+UNet(f); a fixed
state-of-ther-art UNet ISP trained for dark environments \cite{punnappurath2022day}
is used following auto-white balance (AWB) and simple gamma (sGM).
The AWB and sGM are identical to the preprocessing pipeline of the
trained UNet. In this setting, the UNet ISP is trained with paired
data for dark environments proposed in \cite{punnappurath2022day}.
The third is AW+sGM+UNet(p); end-to-end training is performed from
the pre-trained UNet ISP's weight to tune it to the target sensor
and downstream detector. 

The other category is classical function-based ISPs. As for NeuralAE,
we do not convert the images to those taken with lower-quality sensors
as the original paper did to make a fair comparison and use our ISP
functions because the details are missing. Comparing DNN ISP and classical
ISPs, the tuned classical ISPs outperform the existing state-of-the-art
DNN ISPs in challenging environments despite the low computational
cost. One of the reasons should be the lack of paired data for the
target dataset to train the DNN ISPs, but it is difficult to create
ground truth images for challenging RAW data. ISPs that do not require
paired data and can be optimized with downstream task loss must be
useful. As to NeuralAE \cite{onzon2021neural}, we try two of its
proposed encoders, \emph{Semantic Feature Branch} and \emph{Global
Feature Branch}. Our \emph{Global Feature Branch} implementation uses
the histc() function implemented in Pytroch \cite{paszke2019pytorch}
to generate multiple histograms of the image, but it is time-consuming.
So we only use \emph{Semantic Feature Branch} in our method. We also
tried further tuning from the well-trained diff. tuning (GM) \cite{onzon2021neural}
weight using Hardware-in-the-loop \cite{yoshimura2022RAWAug}, but
it did not improve accuracy due to overfitting the training data.
Although we use vanilla CMA-ES in the Hardware-in-the-loop due to
the missing information about the modifications, it should be no problem
since only three parameters in the GM function are the target for
the optimization.

Comparing ours with both categories, our method achieves the best
accuracy in spite of the low computational cost. Especially, compared
with diff. tuning, ours further improved detection accuracy with a
little computational gain. While our computational cost and inference
speed are almost unchanged compared to static ISP (diff. tuning),
the number of parameters is increased. This is because the fully connected
layers whose channel sizes are 1024 in the \emph{Semantic Feature
Branch} have many parameters despite a few computational costs. If
the target device is constrained by the number of parameters, we can
reduce their channel sizes. It is not a bottleneck, but an enough
large value of 256 is also set for the channel size for latent variable
in LU since the computational cost hardly increases. Although NeuralAE
also controls the exposure function, it is a simple function. Instead,
ours enables the control of complicated functions and yields more
suitable images for the downstream detector like Fig. \ref{fig:vizu}.

\subsection{Low Light Image Recognition}

We evaluate on the public LODDataset \cite{Hong2021Crafting}. The
experimental settings are almost identical to \cite{Hong2021Crafting}
using CenterNet \cite{zhou2019objects} as a detector. The only one
difference is that we apply wider contrast augmentation from 1/100
to 1/10 since the dataset is captured with 1/100 to 1/10 times of
default ISO and our input are not normalized with the ISO factor.
Following \cite{Hong2021Crafting}, we use simulated RAW-like images
converted from COCO Dataset \cite{lin2014microsoft} using UPI \cite{brooks2019unprocessing}
as a training dataset and use real RAW images as a test dataset. AG,
GM, and CS are used for the ISP as this combination is deemed effective
for LODDataset. It might be because this dataset is not so much darker
than the darkest image in the human detection dataset.

As shown in Table \ref{tab:coco2lod}, our method achieves significant
improvement from the state-of-the-arts. The dynamic control may mitigate
the domain gap between real RAW images and simulated RAW-like images.
Because the control is based on what the detector felt and reacted
to the input image, it automatically self-adjusts and closes the domain
gap. Our method outperformed NeuralAE by a large margin by enabling
the control of complex ISP functions. More results are listed in the
supplemental material.

\begin{table}[h]
\caption{Evaluation on LODDataset trained with simulated RAW-like data converted
from COCO Dataset \cite{lin2014microsoft}. Note that NeuralAE and
ours are compared with the same ISP combinations. NeuralAE always
applies AE outside ISPs, but it works samely with AG on the software
computation. }

\begin{center}

\scalebox{0.9}{

\begin{tabular}{c|c}
\hline 
 & AP@0.5:0.95\tabularnewline
\hline 
\hline 
as is & 14.3\tabularnewline
SID \cite{chen2018learning} & 16.2\tabularnewline
Zero DCE \cite{guo2020zero} & 21.6\tabularnewline
REDI \cite{lamba2021restoring} & 25.4\tabularnewline
H. Yang et. al. \cite{Hong2021Crafting} & 28.8\tabularnewline
diff. tuning (ISP=GM) \cite{yoshimura2022RAWAug} & 33.0\tabularnewline
NeuralAE (ISP=GM)\cite{onzon2021neural} & 33.7\tabularnewline
NeuralAE (ISP=GM+CS)\cite{onzon2021neural} & 30.6\tabularnewline
ours (ISP=GM) & 43.6\tabularnewline
ours (ISP=AG+GM+CS) & \textbf{44.0}\tabularnewline
\hline 
\end{tabular}}

\end{center}

\vspace{-6mm}

\label{tab:coco2lod}
\end{table}

\section{Conclusion}

We propose DynamicISP, a method to dynamically control parameters
of the ISP functions per image, which is beneficial to recognition
in various environments. Although they are classical functions, the
dynamic control compensates expressive power. Our controller is based
on how the downstream recognition model reacted and desired at the
previous time step. The control is realized with the proposed \emph{residual
output format of parameters}, \emph{latent update style controller},
and \emph{parameter initializer}. Finally, we achieve state-of-the-art
detection results with efficient computing costs for detection in
various environments. We believe DynamicISP will enhance many other
computer vision tasks, such as image classification, keypoint detection,
and segmentation. Furthermore, the proposed controller might enhance
other modal tasks like sound recognition and natural language processing.


{\small
\bibliographystyle{ieee_fullname}
\bibliography{DynamicISP_arxiv_ver7.bbl} 
}{\small\par}
\appendix

\part*{Appendices}

\section{Details of the Implemented ISP}

We implement the following ISP functions in a differentiable manner.

\subsubsection*{Auto Gain (AG)}

\begin{figure}
\centering

\includegraphics[width=0.75\linewidth]{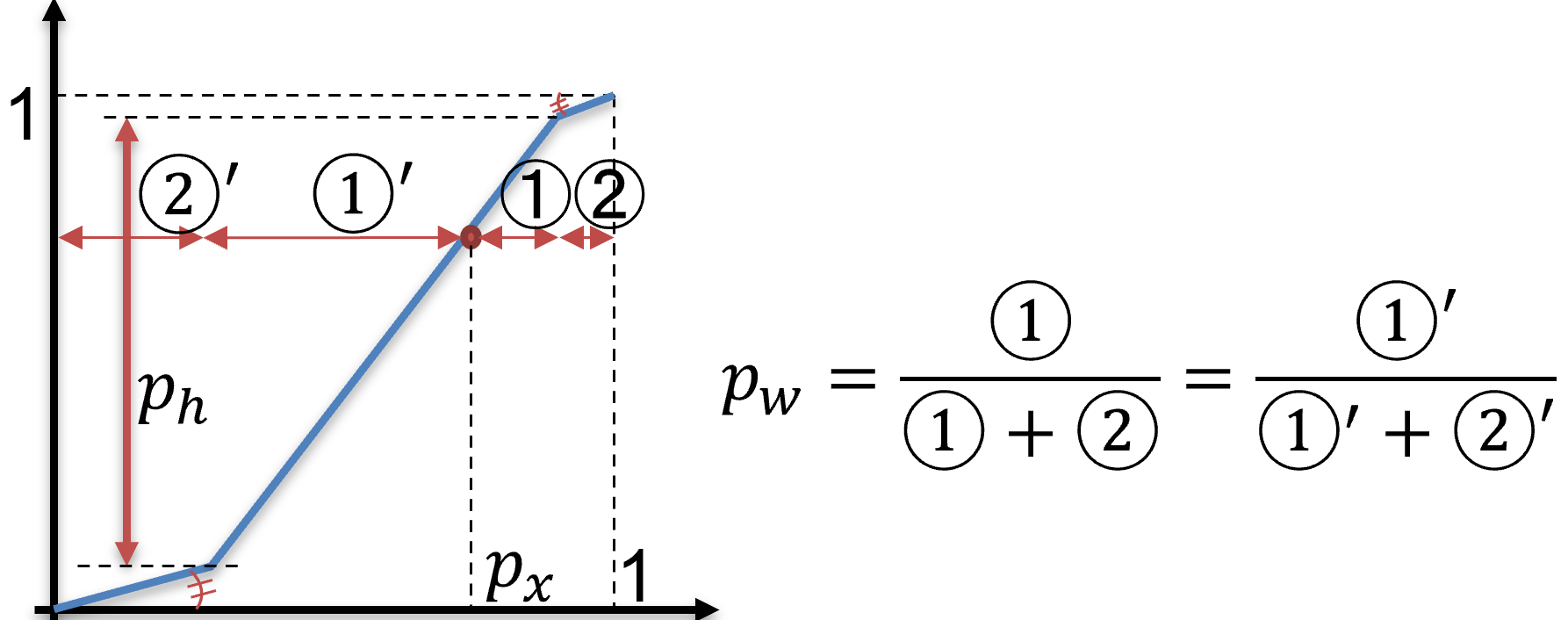}

\caption{Auto gain function without clipping.}

\label{fig:gain}

\vspace{-4mm}
\end{figure}

Usual auto gains simply multiply some value, but we formulate as,

\begin{equation}
I_{AG}\left(X\right)=\begin{cases}
\frac{1-p_{h}}{1-p_{w}}X\\
\;\left(if\,x<p_{x}\left(1-p_{w}\right),\,x\in X\right)\\
\frac{1-p_{h}}{1-p_{w}}X+\frac{p_{h}-p_{w}}{1-p_{w}}\\
\;\left(if\,p_{x}\left(1-p_{w}\right)+p_{w}<x,x\in X\right)\\
\frac{p_{h}}{p_{w}}(X-p_{x}(1-p_{w}))+p_{x}(1-p_{h})\\
\;\left(otherwise\right)
\end{cases},\label{eq:ag}
\end{equation}
 where $p_{w}$, $p_{h}$, and $p_{x}$ are parameters to be controlled,
and $x$ is the each pixel in the input image $X$. It intends to
control how much and what range of domain should be emphasized with
the third equation. The first and the second avoid overflow from $[0,1]$
without clipping as shown in Fig. \ref{fig:gain}.

\subsubsection*{Denoiser (DN)}

We utilize a simple Bilateral filter (BF) \cite{tomasi1998bilateral}
as, 
\begin{equation}
I_{DN}\left(X\right)=\left(1-p_{a}\right)\cdot X+p_{a}\cdot BF\left(p_{\sigma s},p_{\sigma i};X\right),\label{eq:dn}
\end{equation}
where $p_{\sigma s}$ and $p_{\sigma i}$ are the parameters for the
spatial and intensity variance and $p_{a}$ is another parameter.
We set the kernel size as five.

\subsubsection*{Sharpener (SN)}

Simple Gaussian filter (GF) is used as, 
\begin{equation}
I_{SN}\left(X\right)=\left(1-p_{a}\right)\cdot X+p_{a}\cdot\left(X-GF\left(p_{\sigma};X\right)\right).\label{eq:sn}
\end{equation}
 The second term is the difference-of-Gaussians \cite{marr1980theory}
whose kernel sizes are one and five;

\begin{align}
DoG(X) & =GF(p_{\sigma1},k_{1};X)-GF(p_{\sigma},k_{2};X)\nonumber \\
 & =X-GF(p_{\sigma},5;X).\label{eq:10}
\end{align}

\subsubsection*{Gamma (GM)}

We follow the parameterization of gamma tone mapping in \cite{mosleh2020hardware,yoshimura2022RAWAug}
and implement it as differentiable; 
\begin{equation}
I_{GM}\left(X\right)=X^{\frac{1}{p_{\text{\textramshorns1}}}\cdot\frac{1-\left(1-p_{\text{\textramshorns2}}\right)X^{\frac{1}{p_{\text{\textramshorns1}}}}}{1-\left(1-p_{\text{\textramshorns2}}\right)p_{k}^{\frac{1}{\text{\ensuremath{p_{\text{\textramshorns1}}}}}}}}.\label{eq:gm}
\end{equation}

\subsubsection*{Contrast Stretcher (CS)}

We implement CS as a simple linear function of $I_{CS}\left(X\right)=q_{b}X+q_{c}$.
Because DNN can process any range of value, we do not restrict the
range.

\section{Additional Evaluations}

\subsection{Evaluation on Human Detection}

\subsubsection*{Multi-Layer Control}

A more detailed ablation study is performed for multi-layer control.
Here, we add ISP layers in order of effect one by one. As the Table
\ref{tab:multi_layer_val} shows, the proposed method without LU struggles
to control multi-layer ISPs. The proposed LU successfully disentangles
the difficulty of multi-layer control and boosts the accuracy from
the setting of only containing GM tone mapping. 

\begin{table}[h]
\caption{Control of Multi-layer ISPs on the human detection dataset. We add
ISP layers in order of effect.}

\begin{center}

\begin{tabular}{c|cc}
\hline 
ISP components & w/o LU & w/ LU\tabularnewline
\hline 
\hline 
GM & 48.9 & -\tabularnewline
GM+CS & 49.4 & 49.4\tabularnewline
DN+GM+CS & 49.2 & 49.4\tabularnewline
DN+SN+GM+CS & 48.0 & \textbf{49.5}\tabularnewline
\hline 
\end{tabular}

\end{center}

\label{tab:multi_layer_val}
\end{table}

\subsubsection*{Comparison with Other Possible Controllers}

\begin{figure*}
\centering

\begin{tabular}{cc}
\def\svgwidth{1.0\columnwidth}\scriptsize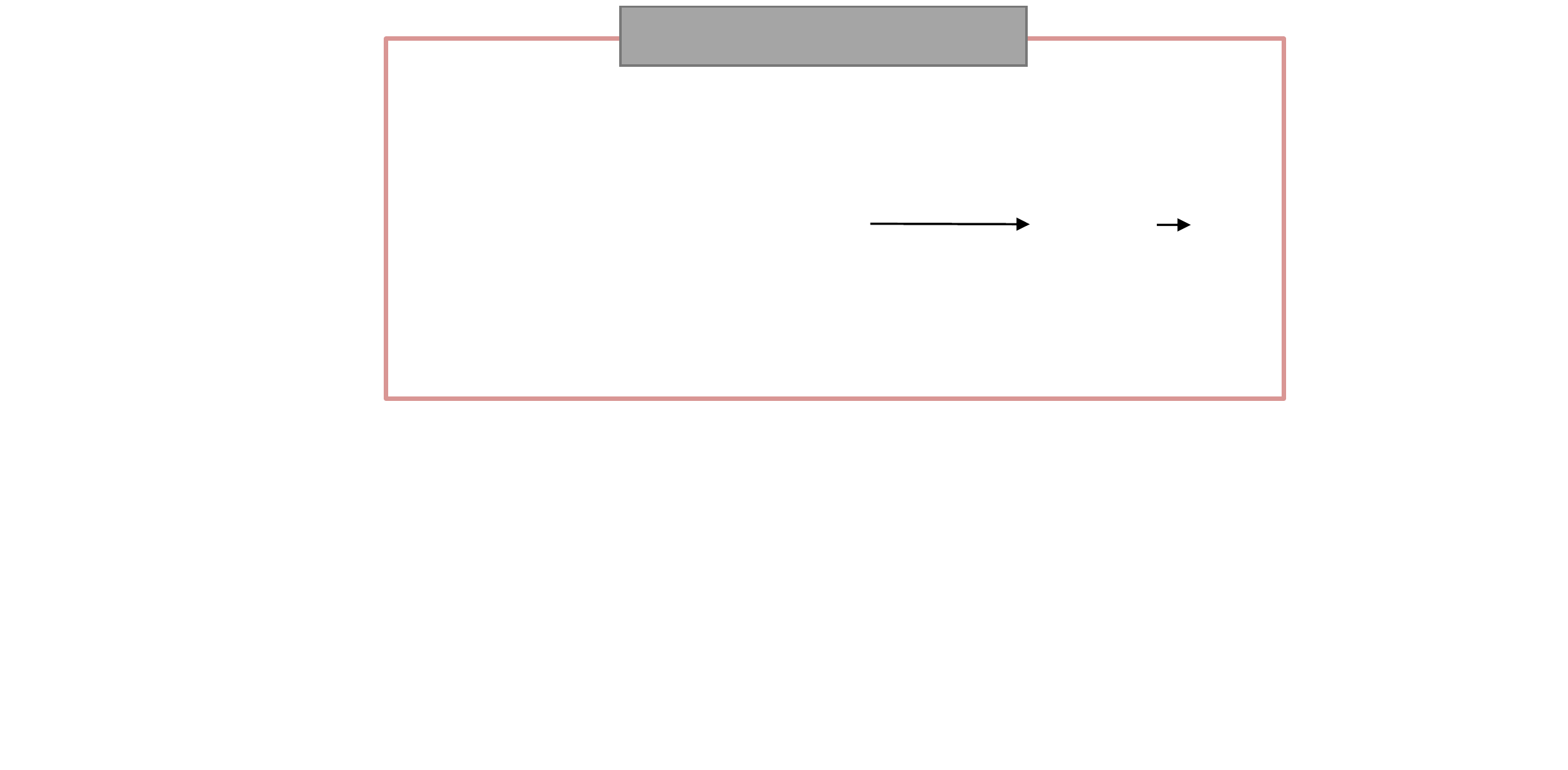 & \def\svgwidth{1.0\columnwidth}\scriptsize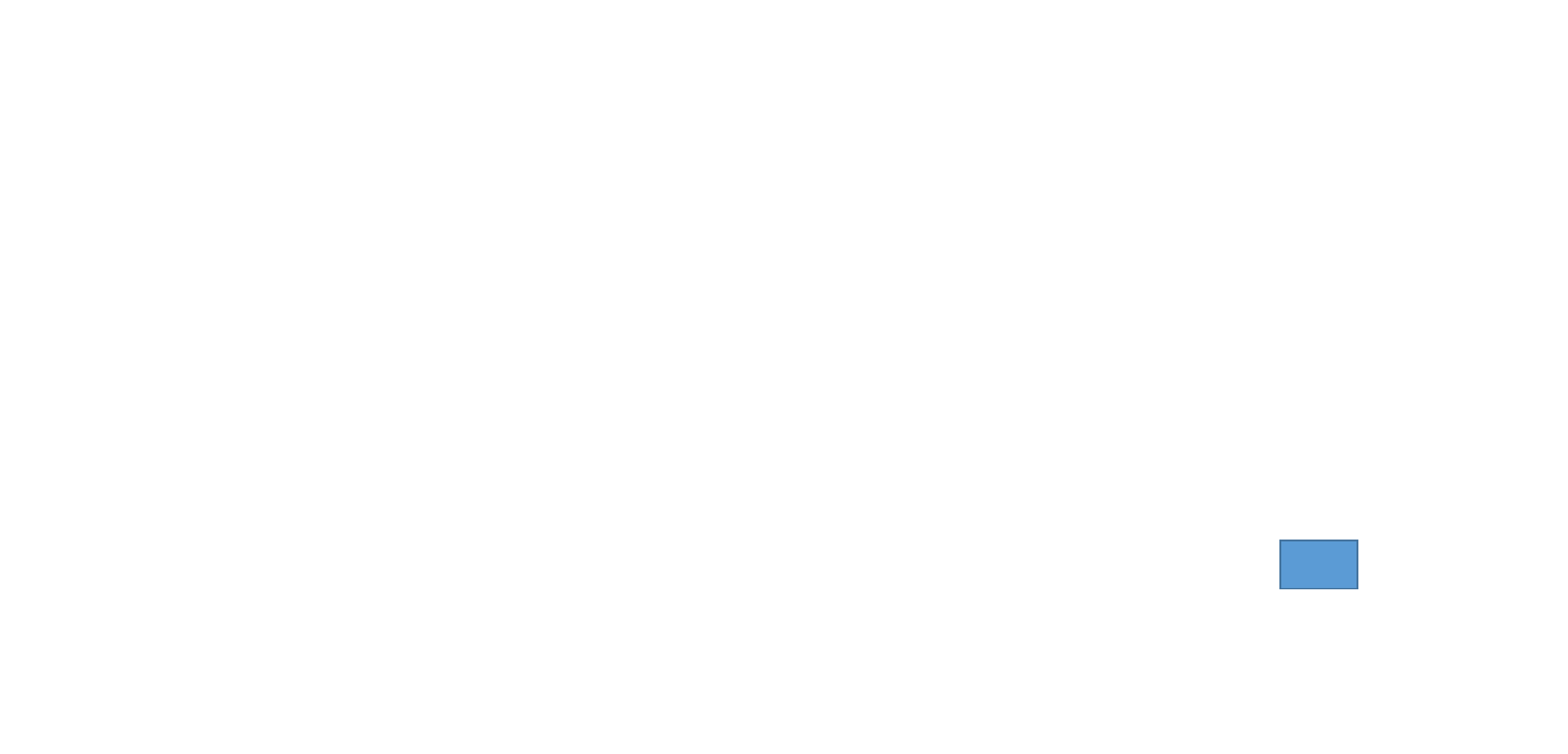\tabularnewline
\scalebox{0.8}{(a) Feedforward control from a single-feature.} & \scalebox{0.8}{(b) Feedforward control from multi-features.}\tabularnewline
 & \tabularnewline
\def\svgwidth{1.0\columnwidth}\scriptsize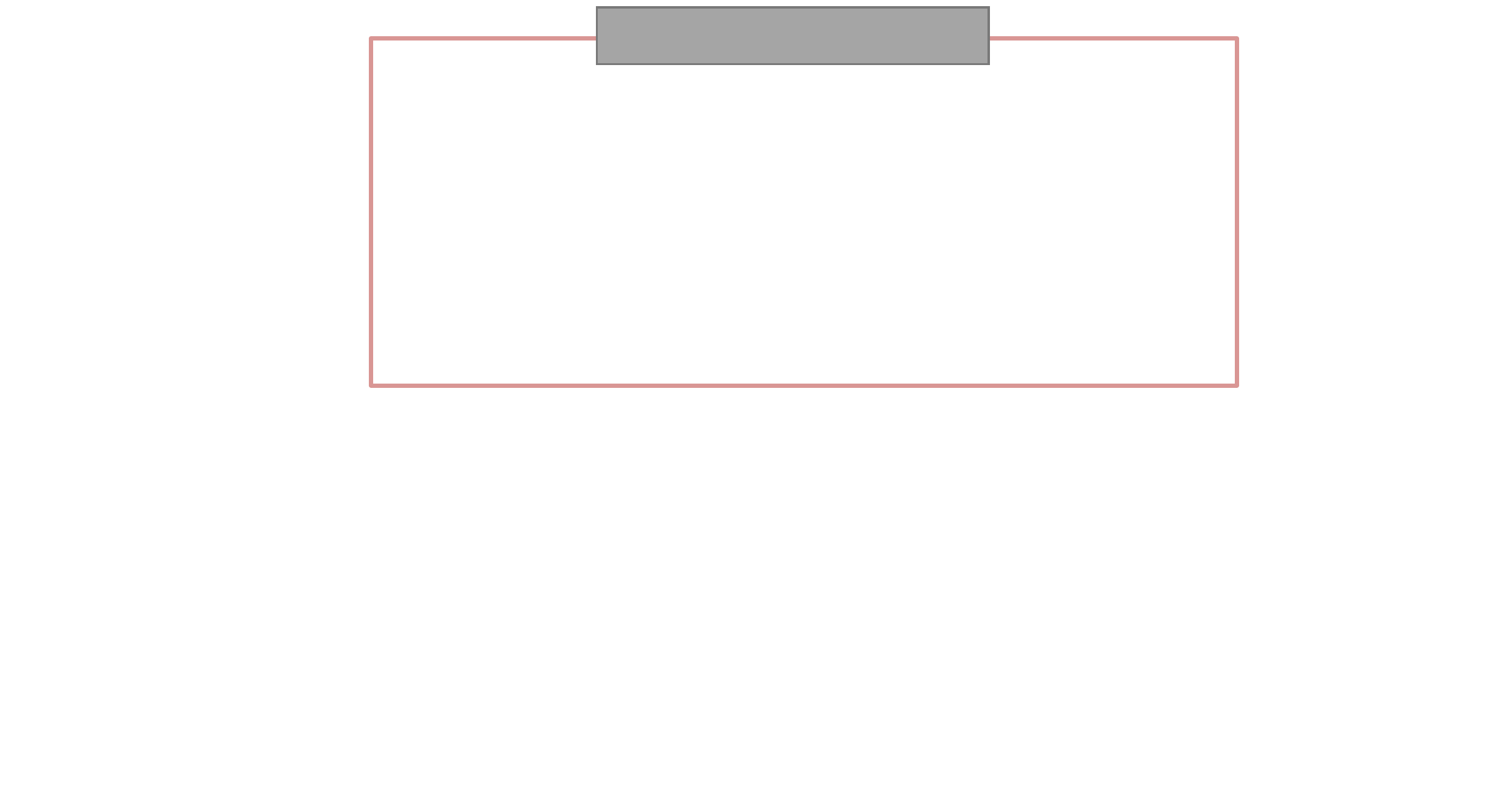 & \def\svgwidth{1.0\columnwidth}\scriptsize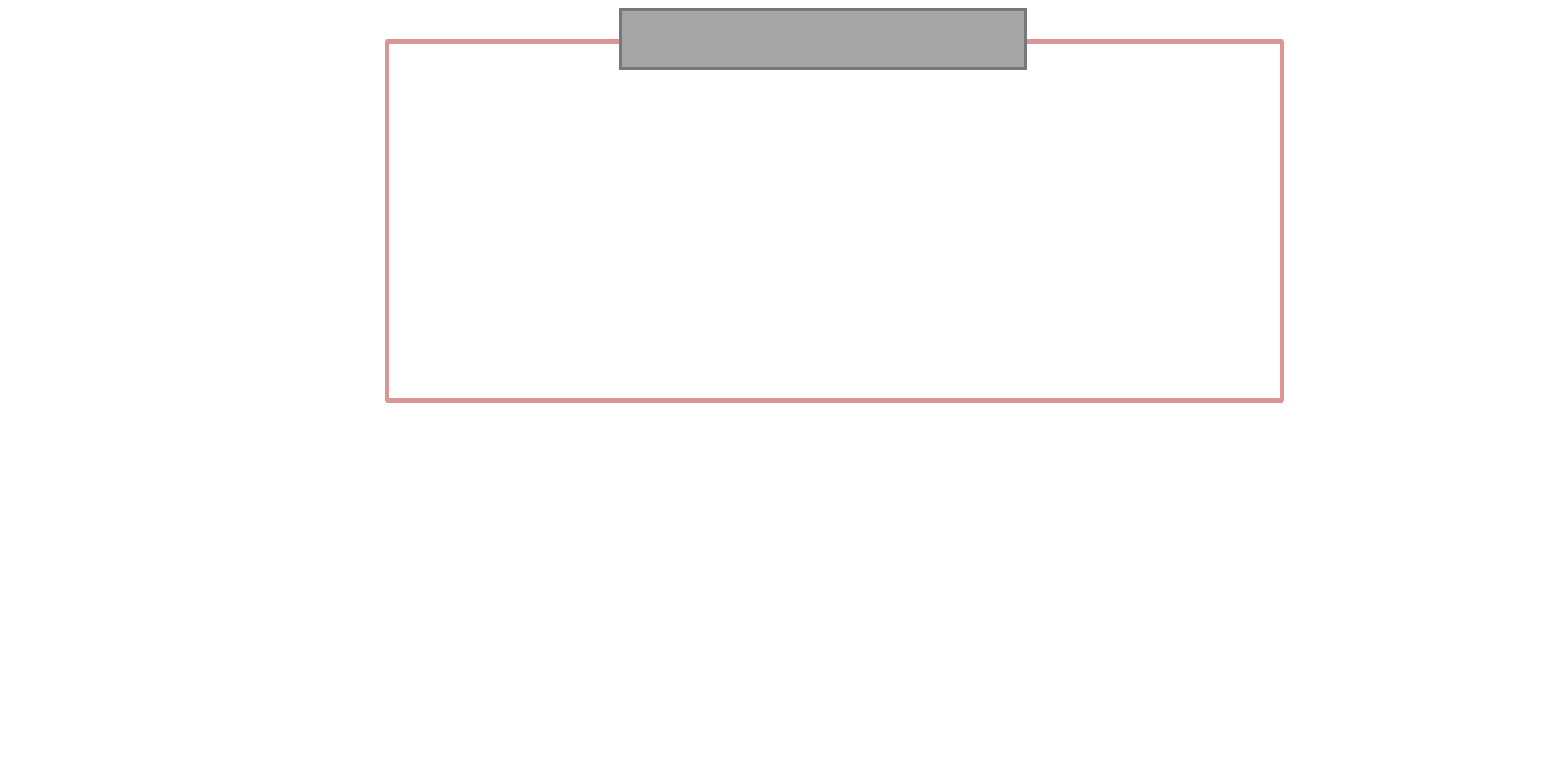\tabularnewline
\scalebox{0.8}{(c) Feedforward control by exemplifications} \\
\scalebox{0.8}{when using several ISP parameter sets \cite{conde2022model}.} & \scalebox{0.8}{(d) Feedback control from a single-feature (ours).}
\\ \scalebox{0.8}{}\tabularnewline
\end{tabular}

\caption{Comparison with other possible controllers. The (a) is the most lightweight
possible controller among feedforward control. The (b) is based on
typical dynamic neural network architectures \cite{yang2019condconv,chen2020dynamicc,chen2020dynamic}.
The (c) is a more advanced method of the (b) proposed for RAW image
reconstruction \cite{conde2022model}. The (d) is the proposed efficient
feedback control.}

\label{fig:arch_comparison}
\end{figure*}

\begin{table*}
\caption{Comparison with other possible controllers on the human detection
dataset. $C$ is the computational cost of the ISP. ResNet18 or 4-layer
CNN is used as the encoder for feedforward controls.}

\begin{center}

\scalebox{0.9}{

\begin{tabular}{cc|ccccc}
\hline 
ISP &  & GFLOPS & \#params {[}M{]} & latency {[}ms{]} & controller {[}ms{]} & AP {[}\%{]}\tabularnewline
\hline 
\hline 
\multirow{5}{*}{GM} & (a) (encoder = 4-layer CNN) & 14.03+C & \textbf{14.33} & 11.2 & 3.3 & 47.5\tabularnewline
 & + our ``RO+'' & 14.03+C & \textbf{14.33} & 11.5 & 3.6 & 48.8\tabularnewline
 & (a) (encoder = ResNet18) & 20.99+C & 25.77 & 13.2 & 5.3 & 46.1\tabularnewline
 & + our ``RO+'' & 20.99+C & 25.77 & 13.6 & 5.7 & \textbf{49.4}\tabularnewline
 & (d) ours & \textbf{13.65+C} & 15.66 & \textbf{8.3} & \textbf{0.4} & 49.1\tabularnewline
\hline 
\multirow{6}{*}{DN+SN+GM+CS} & (a) (encoder = ResNet18) & 20.99+C & 25.77 & 31.1$\spadesuit$ & 5.3 & 48.3\tabularnewline
 & + our ``RO+'' and ``LU'' & 20.99+C & 25.87 & 31.6$\spadesuit$ & 5.8 & 48.5\tabularnewline
 & (b) (encoder = 4-layer CNN) & 15.23+C & \textbf{14.72} & 29.1$\spadesuit$ & 3.3 & 47.7\tabularnewline
 & + our ``RO+'' & 15.23+C & \textbf{14.72} & 29.5$\spadesuit$ & 3.7 & 49.0\tabularnewline
 & (c) (encoder = 4-layer CNN) & 15.93+6C & 14.73 & 101.1$\spadesuit$ & 9.2 $\clubsuit$ & 49.4\tabularnewline
 & (d) ours & \textbf{13.65+C} & 15.76 & \textbf{27.7}$\spadesuit$ & \textbf{1.9} & \textbf{49.6}\tabularnewline
\hline 
\end{tabular}}

\end{center}

\scalebox{0.8}{

$\spadesuit$: Our implementations of DN and SN have a lot of duplicate
calculations and are not optimized.

}

\scalebox{0.8}{

$\clubsuit$: Because of the above, the controller's latency is measured
in case all four layers are GM to make a fair comparison.

}

\label{tab:comp_arch}
\end{table*}

\begin{table*}
\caption{Detailed evaluation on LODDataset \cite{Hong2021Crafting} trained
with simulated RAW-like data converted from COCO Dataset \cite{lin2014microsoft}.}

\vspace{4mm}

\begin{center}

\begin{tabular}{c|cccccc|c}
\hline 
 & \multicolumn{7}{c}{mAP@0.5:0.95 per exposure ratio to default}\tabularnewline
 & 1/10 & 1/20 & 1/30 & 1/40 & 1/50 & 1/100 & Ave.\tabularnewline
\hline 
\hline 
as is & 23.3 & 16.7 & 14.4 & 114.1 & 13.4 & 3.6 & 14.3\tabularnewline
SID \cite{chen2018learning} & 25.8 & 20.0 & 16.4 & 15.1 & 13.2 & 6.7 & 16.2\tabularnewline
Zero DCE \cite{guo2020zero} & 32.5 & 25.3 & 23.4 & 21.5 & 17.8 & 8.9 & 21.6\tabularnewline
REDI \cite{lamba2021restoring} & 33.6 & 30.2 & 26.1 & 24.6 & 23.4 & 14.1 & 25.4\tabularnewline
H. Yang et. al. \cite{Hong2021Crafting} & 38.5 & 31.7 & 29.3 & 27.8 & 27.1 & 18.1 & 28.8\tabularnewline
diff. tuning (GM) \cite{yoshimura2022RAWAug} & 42.8 & 34.9 & 39.5 & 38.9 & 29.4 & 12.6 & 33.0\tabularnewline
NeuralAE (GM)\cite{onzon2021neural} & 42.3 & 35.3 & 38.5 & 40.6 & 29.9 & 15.6 & 33.7\tabularnewline
NeuralAE (GM+CS)\cite{onzon2021neural} & 37.0 & 31.5 & 33.4 & 35.1 & 28.5 & 18.3 & 30.6\tabularnewline
ours (GM) & 45.2 & \textbf{41.1} & 51.1 & \textbf{49.1} & \textbf{41.4} & 33.9 & 43.6\tabularnewline
ours (AG+GM+CS) & \textbf{45.8} & \textbf{41.1} & \textbf{52.1} & 48.6 & 41.2 & \textbf{34.9} & \textbf{44.0}\tabularnewline
\hline 
\end{tabular}

\end{center}

\label{tab:coco2lod-1}
\end{table*}

In this section, the proposed controller is compared with other possible
controllers, especially feedforward controllers because most of the
dynamic neural networks methods \cite{yang2019condconv,chen2020dynamicc,chen2020dynamic}
have successfully controlled DNN parameters based on feedforward controls.
Fig. \ref{fig:arch_comparison}(a) controls all functions based on
a single-feature. It is different from the typical controllers for
dynamic neural networks \cite{yang2019condconv,chen2020dynamicc,chen2020dynamic}
but the most lightweight possible feedforward controller. Fig. \ref{fig:arch_comparison}(b)
is based on typical dynamic neural network architectures that control
each layer with an output of the previous layer. The problem in applying
it to the ISP control is that the output of the previous layer is
just an image, so it is necessary to create features from scratch
using an encoder. Fig. \ref{fig:arch_comparison}(c) is a more advanced
method of Fig. \ref{fig:arch_comparison}(b) proposed for RAW image
reconstruction \cite{conde2022model}. Several processed images with
different parameter sets are input to the encoder as exemplifications,
and the output parameters from the encoder are determined as the weighted
average of the parameter sets. Note that the inverse pipeline is not
implemented because it is trained only with detection loss in our
problem setup. The number of exemplifications is set as five. We use
two types of networks with different computational costs as encoders
in Fig. \ref{fig:arch_comparison}(a), (b), and (c): ResNet18 \cite{he2016deep}
or 4-layer light-weight CNNs with ReLU activations, whose kernel sizes,
strides, and output channel sizes are (3, 3, 3, 3), (2, 2, 2, 2),
and (16, 32, 64, 128). Fig. \ref{fig:arch_comparison}(d) is the proposed
feedback control from a feature.

The results are shown in Table \ref{tab:comp_arch}. In the case of
the single-layer ISP setting, the feedforward control exceeds the
accuracy of the proposed method by using a large encoder (ResNet18).
However, in the experimental setting with the 4-layer CNN, where the
computational cost is still higher than the proposed method, the accuracy
is inferior. This result indicates that the feedback control is more
efficient. By adding more convolutions to the ``Semantic Feature
Branch'' (SFB), the proposed feedback control might improve the accuracy.
In our setting, SFB contains only one convolution layer. In addition,
the proposed RO+ for controlling a difficult function is found to
be effective even for the feedforward controls.

In the case of the multi-layer ISP setting, the proposed method outperforms
feedforward controls with lower computational cost. Although the (c)
architecture is accurate, it takes a high computational cost because
it needs multiple computations of ISPs and encoders. Limitted to feedforward
controls, a comparison of (a) and (b) shows that it is more efficient
to encode the previous layer's image with multiple small encoders
than with one large encoder. On the other hand, our feedback control
achieves higher accuracy despite the fact that the control is based
on a single shallow layer of feature (the output of the first stage
of the detector's ResNet18 backbone). This should be because the following
two factors outweigh the difficulty of controlling from a single encoder.
One factor should be the advantage that the controller is able to
extract what is captured by the detector directly. The other factor
should be the effectiveness of the proposed training method for feedback
control (PI).

Lastly, the proposed method is lightweight because it does not require
image encoders and performs almost only 1D tensor operations.

\subsection{Evaluation on Low-Light Recognition}

\subsubsection*{Training with Simulated RAW Images}

\begin{figure*}
\centering

\setlength{\tabcolsep}{0.5pt}


\begin{tabular}{cccc}
\includegraphics[width=0.245\textwidth]{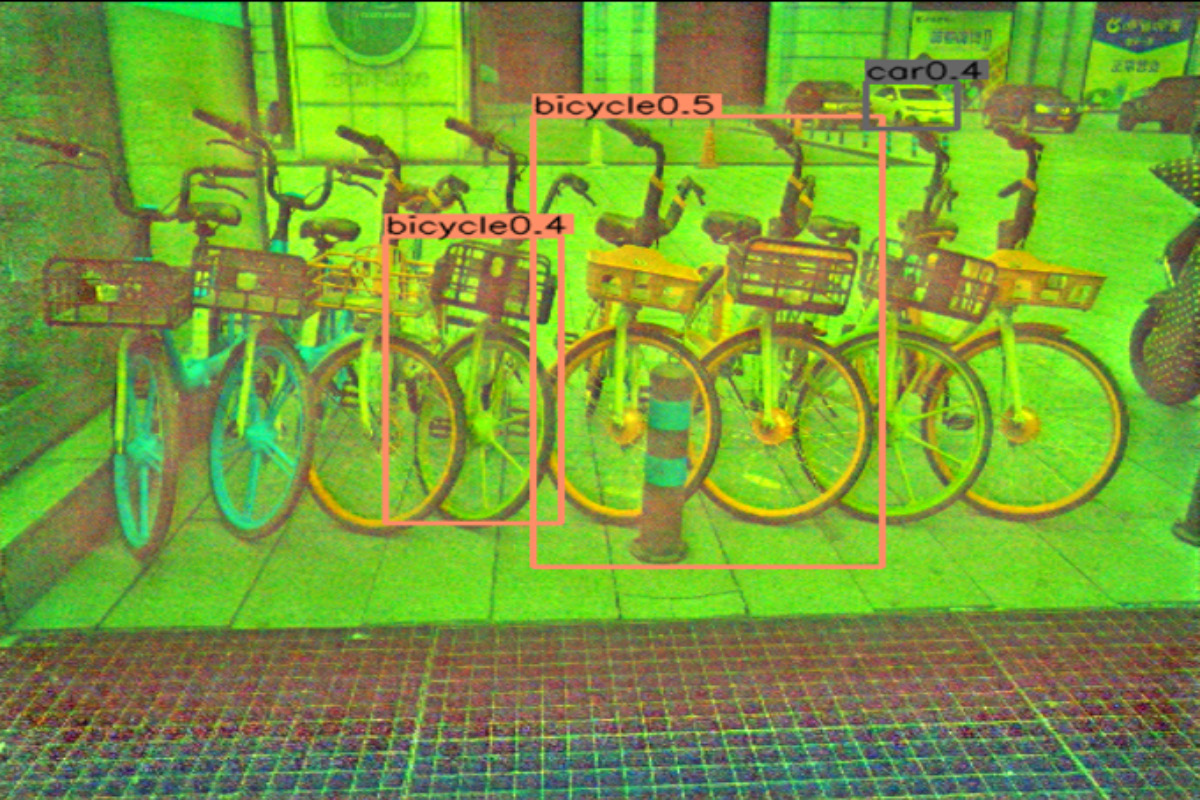} & \includegraphics[width=0.245\textwidth]{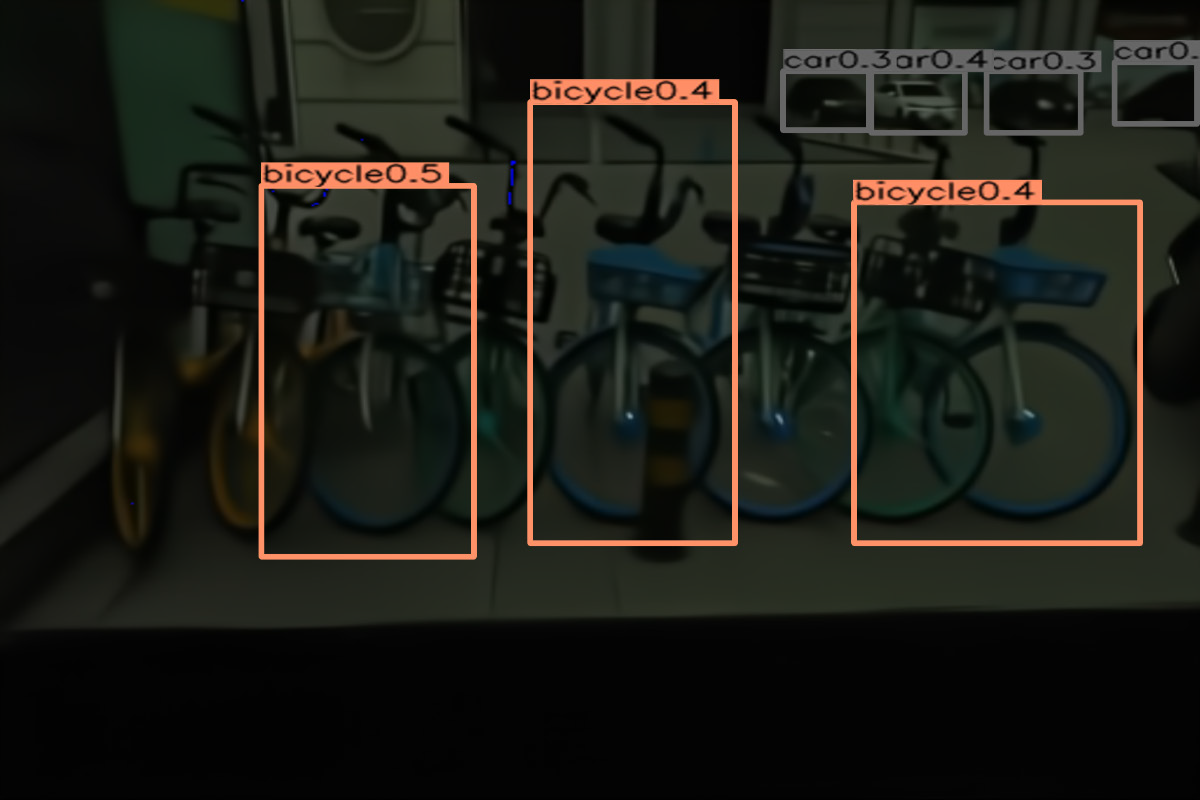} & \includegraphics[width=0.245\textwidth]{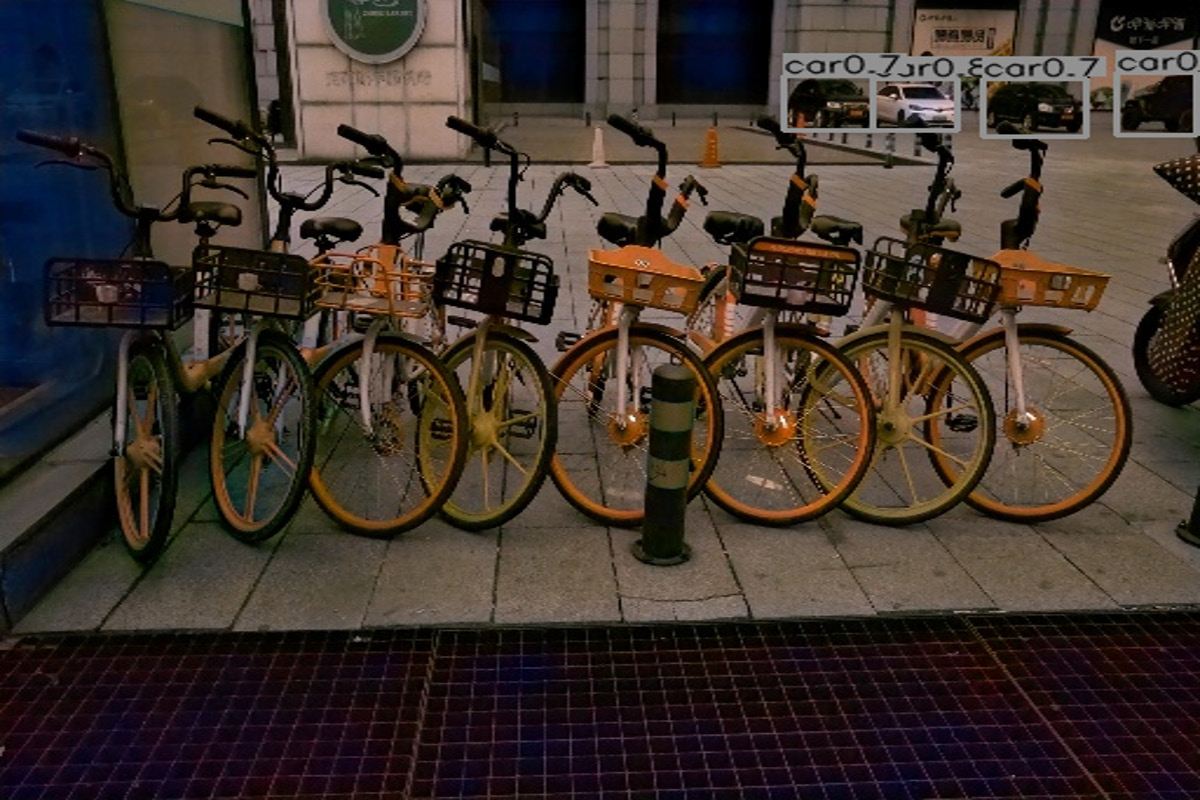} & \includegraphics[width=0.245\textwidth]{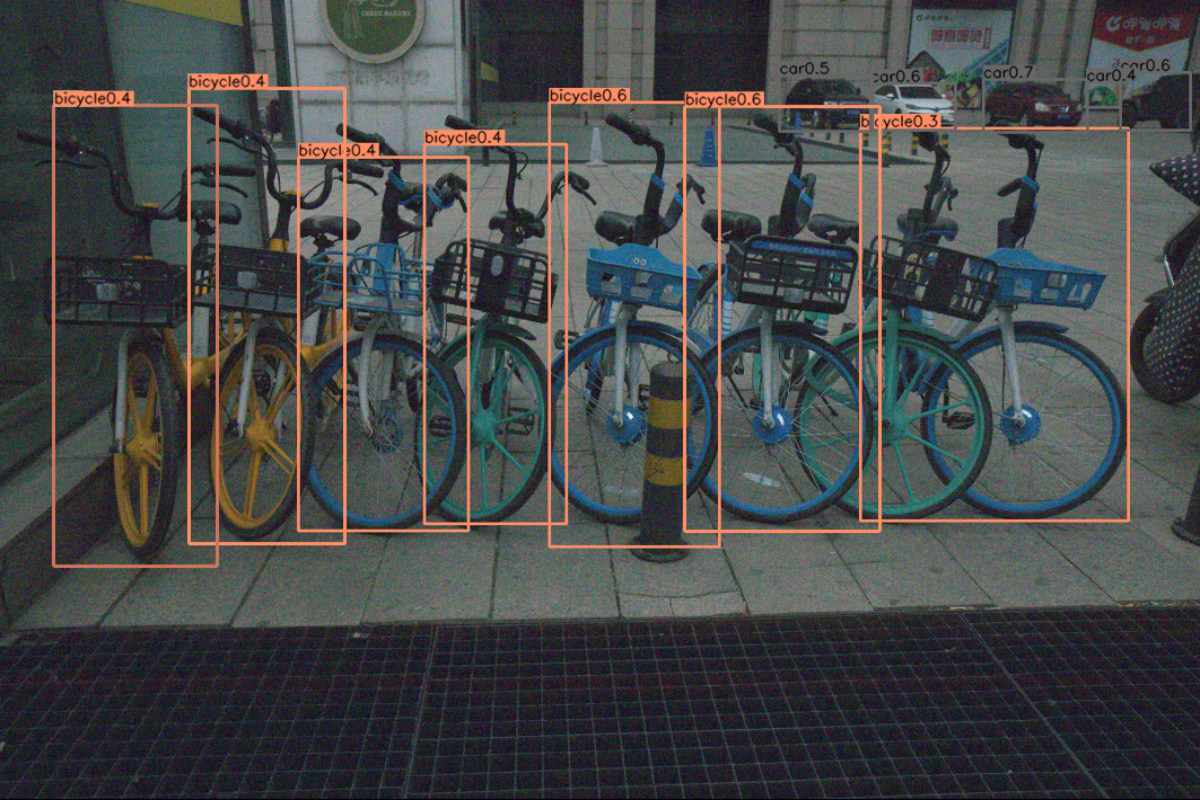}\tabularnewline
\scalebox{0.8}{SID \cite{chen2018learning}} & \scalebox{0.8}{Zero DCE \cite{guo2020zero}} & \scalebox{0.8}{REDI \cite{lamba2021restoring}} & \scalebox{0.8}{H. Yang et. al. \cite{Hong2021Crafting}}\tabularnewline
\includegraphics[width=0.245\textwidth]{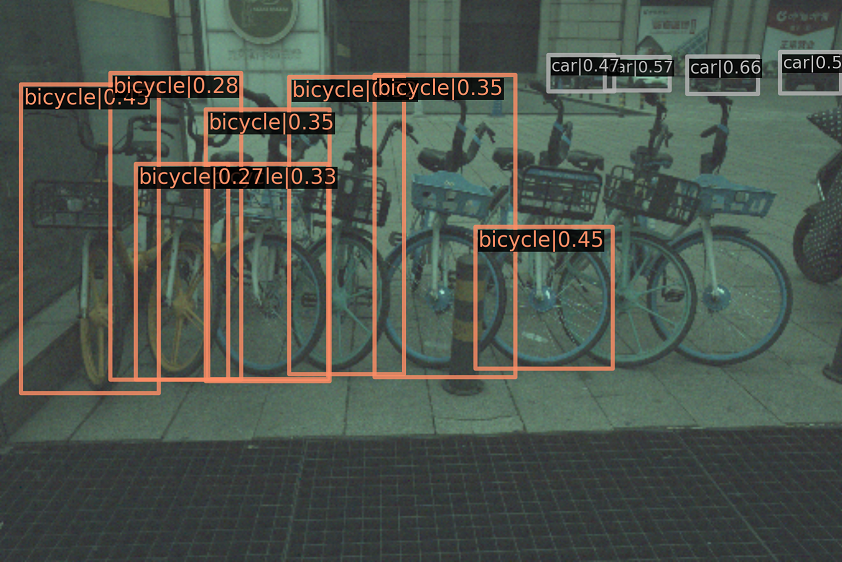} & \includegraphics[width=0.245\textwidth]{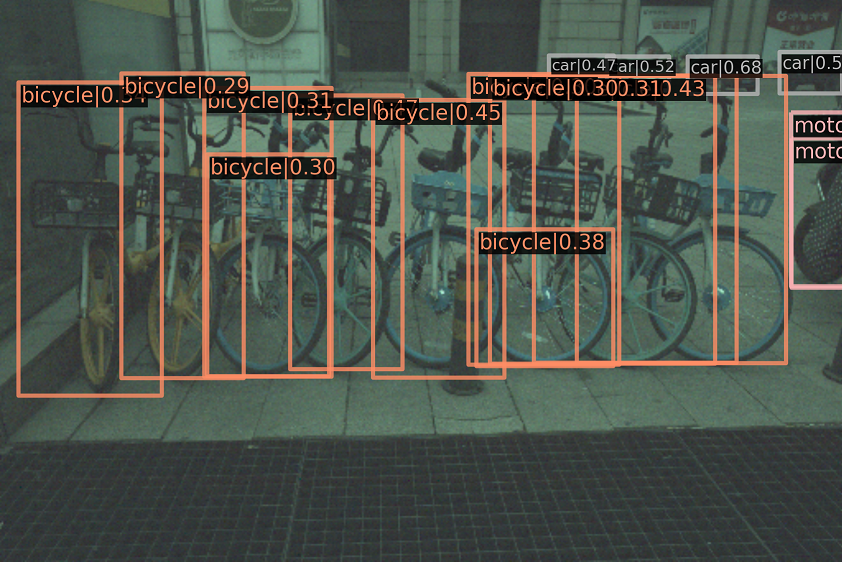} & \includegraphics[width=0.245\textwidth]{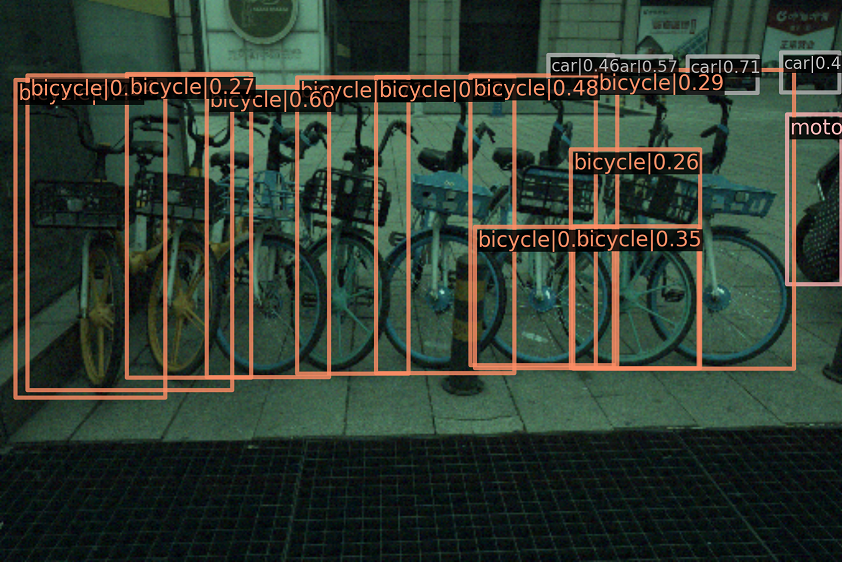} & \includegraphics[width=0.245\textwidth]{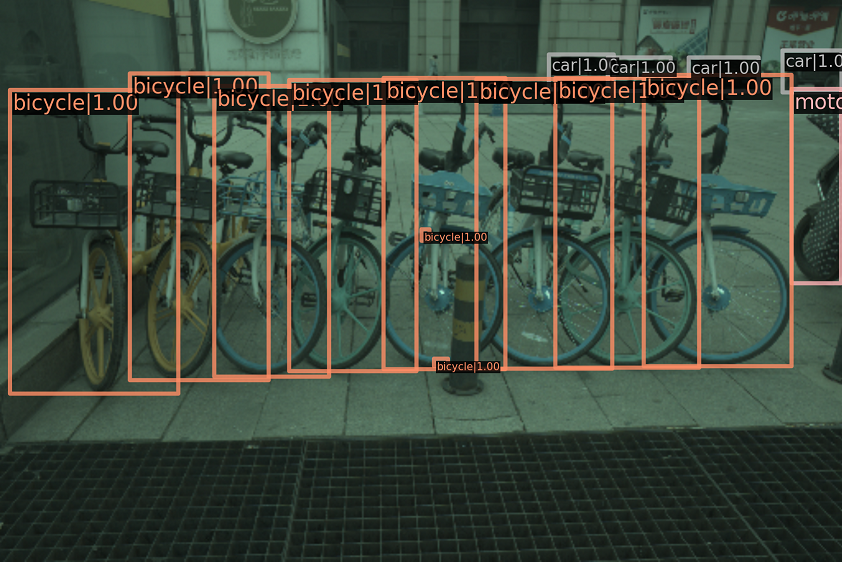}\tabularnewline
\scalebox{0.8}{diff. tuning (GM) \cite{yoshimura2022RAWAug}} & \scalebox{0.8}{NeuralAE (GM)\cite{onzon2021neural}} & \scalebox{0.8}{ours (AG+GM+CS)} & \scalebox{0.8}{GT}\tabularnewline
 &  &  & \tabularnewline
\includegraphics[width=0.245\textwidth]{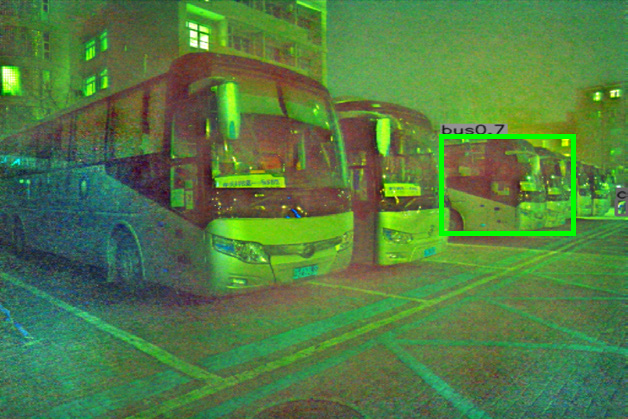} & \includegraphics[width=0.245\textwidth]{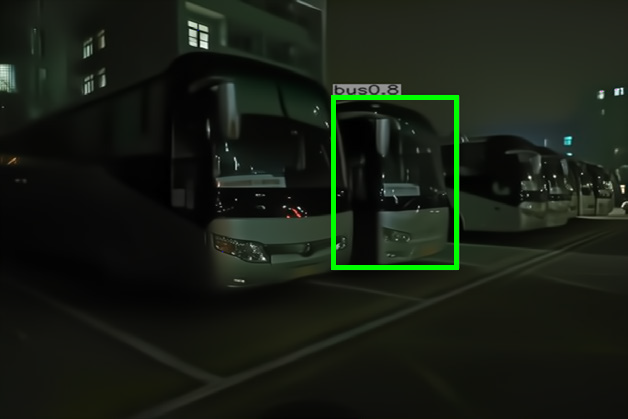} & \includegraphics[width=0.245\textwidth]{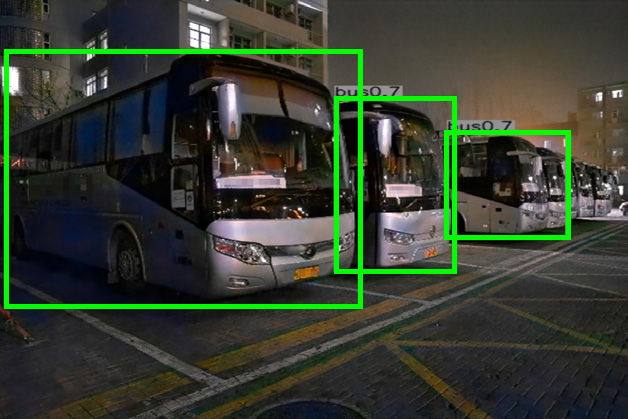} & \includegraphics[width=0.245\textwidth]{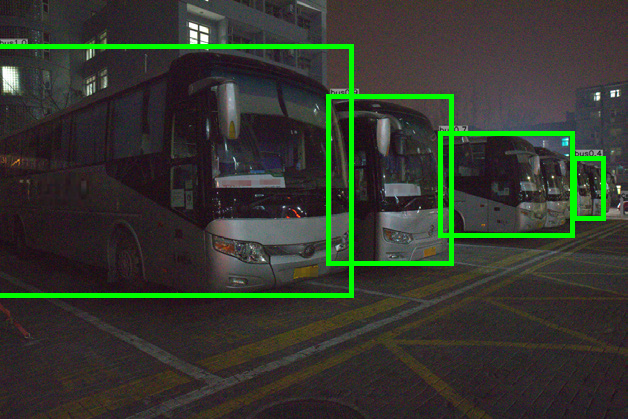}\tabularnewline
\scalebox{0.8}{SID \cite{chen2018learning}} & \scalebox{0.8}{Zero DCE \cite{guo2020zero}} & \scalebox{0.8}{REDI \cite{lamba2021restoring}} & \scalebox{0.8}{H. Yang et. al. \cite{Hong2021Crafting}}\tabularnewline
\includegraphics[width=0.245\textwidth]{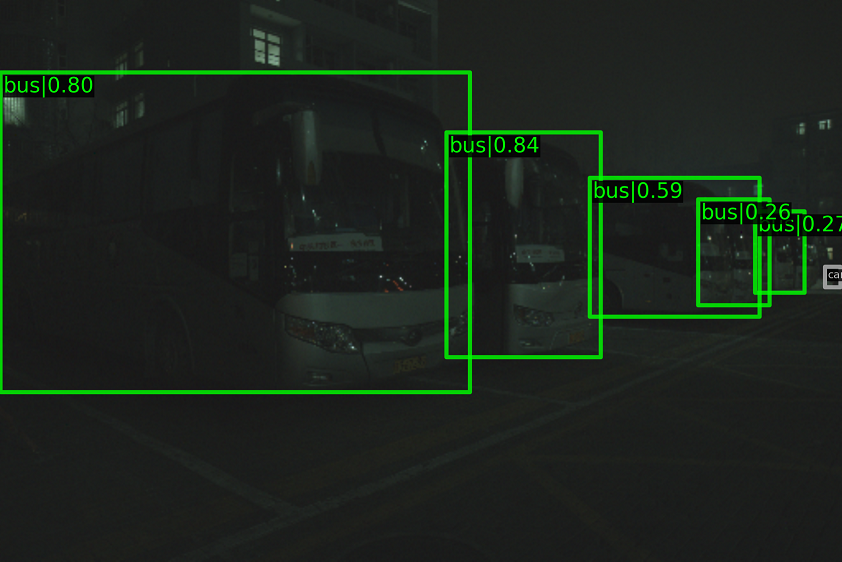} & \includegraphics[width=0.245\textwidth]{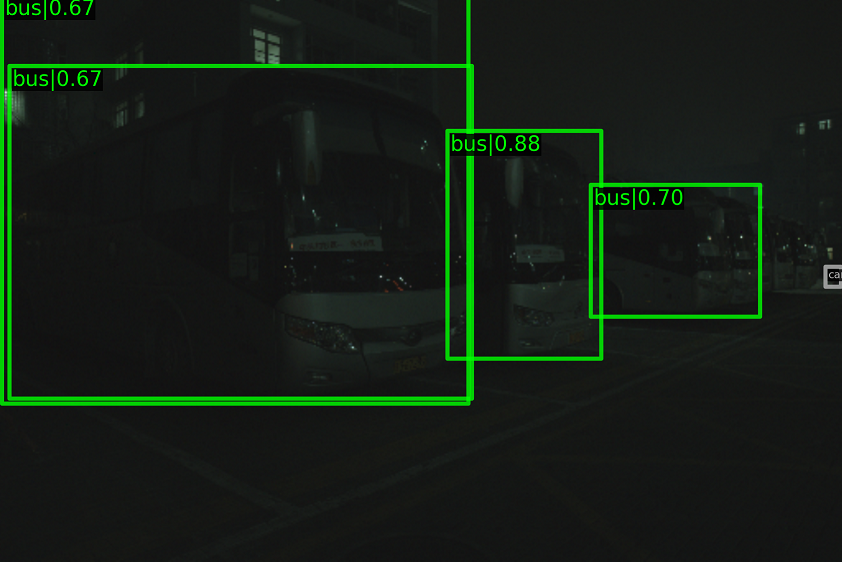} & \includegraphics[width=0.245\textwidth]{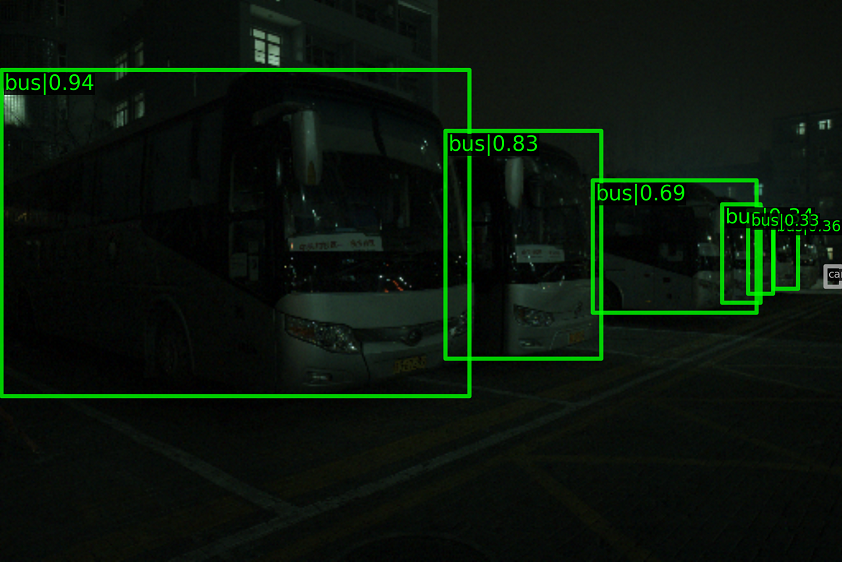} & \includegraphics[width=0.245\textwidth]{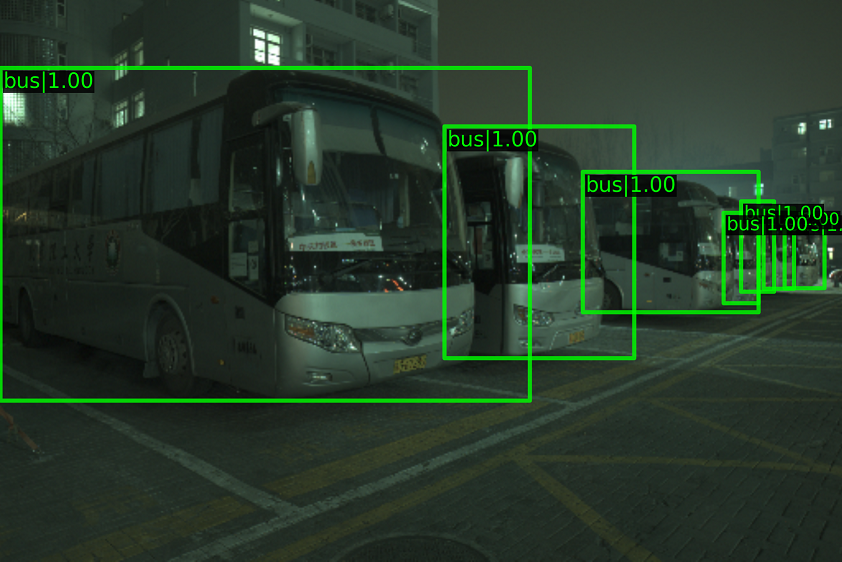}\tabularnewline
\scalebox{0.8}{diff. tuning (GM) \cite{yoshimura2022RAWAug}} & \scalebox{0.8}{NeuralAE (GM)\cite{onzon2021neural}} & \scalebox{0.8}{ours (AG+GM+CS)} & \scalebox{0.8}{GT}\tabularnewline
 &  &  & \tabularnewline
\includegraphics[width=0.245\textwidth]{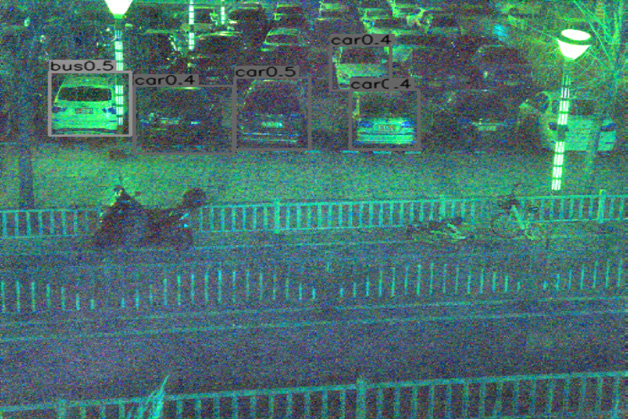} & \includegraphics[width=0.245\textwidth]{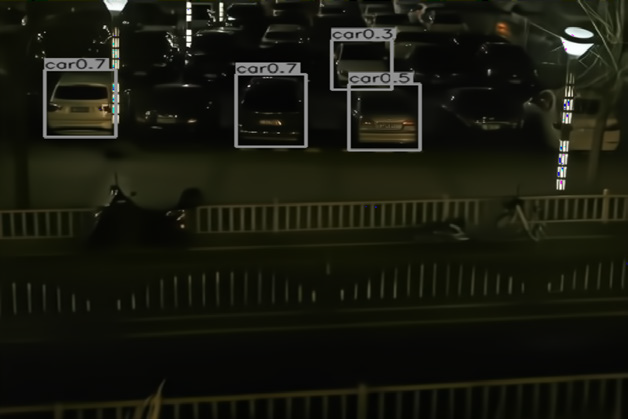} & \includegraphics[width=0.245\textwidth]{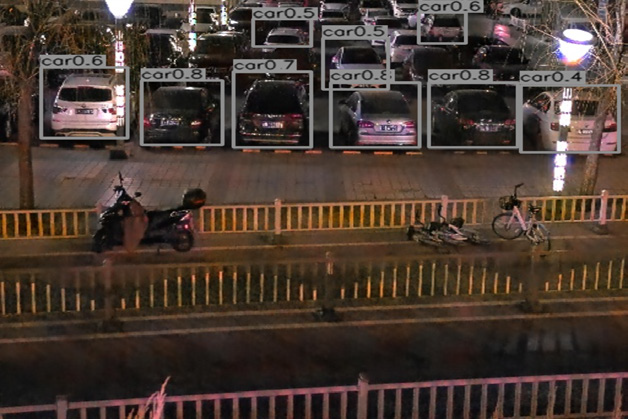} & \includegraphics[width=0.245\textwidth]{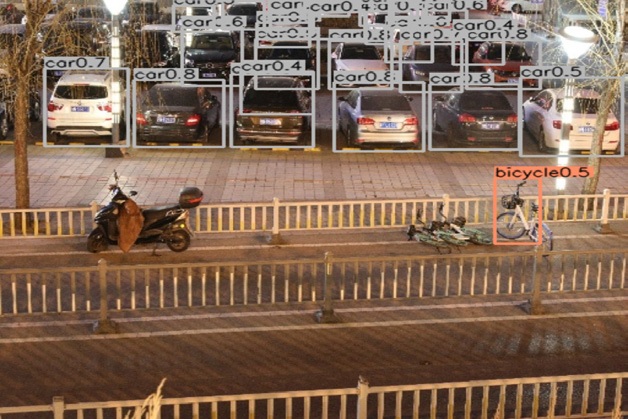}\tabularnewline
\scalebox{0.8}{SID \cite{chen2018learning}} & \scalebox{0.8}{Zero DCE \cite{guo2020zero}} & \scalebox{0.8}{REDI \cite{lamba2021restoring}} & \scalebox{0.8}{H. Yang et. al. \cite{Hong2021Crafting}}\tabularnewline
\includegraphics[width=0.245\textwidth]{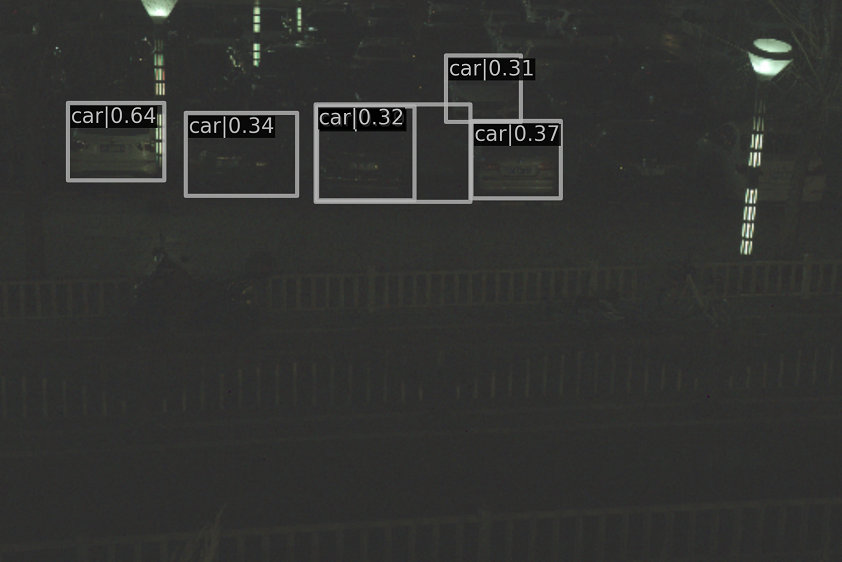} & \includegraphics[width=0.245\textwidth]{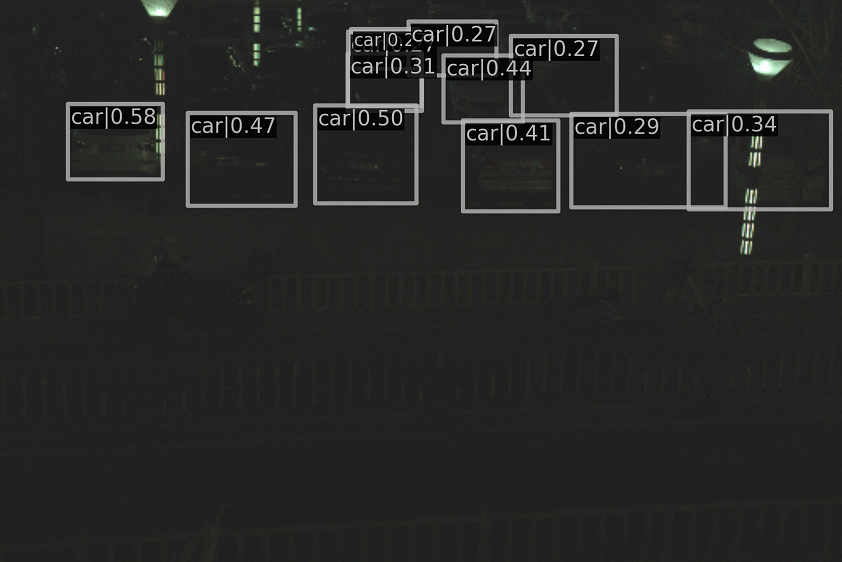} & \includegraphics[width=0.245\textwidth]{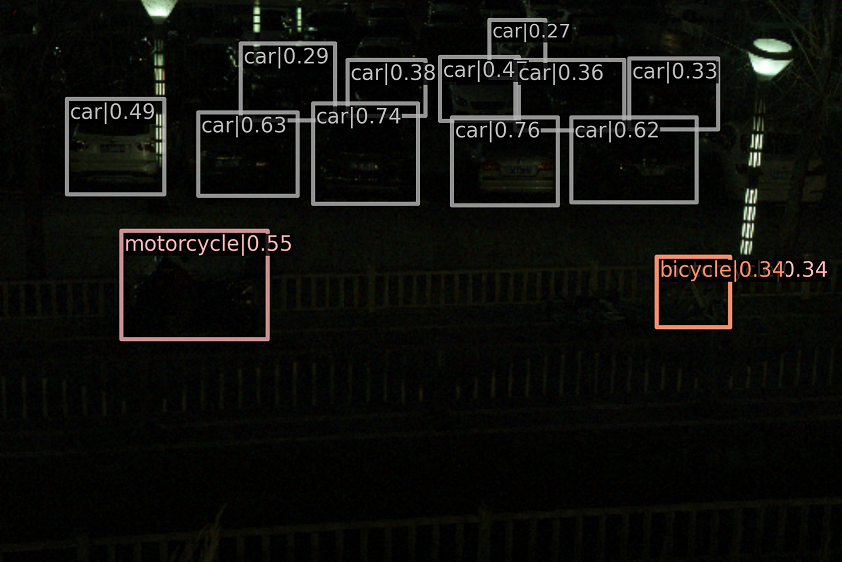} & \includegraphics[width=0.245\textwidth]{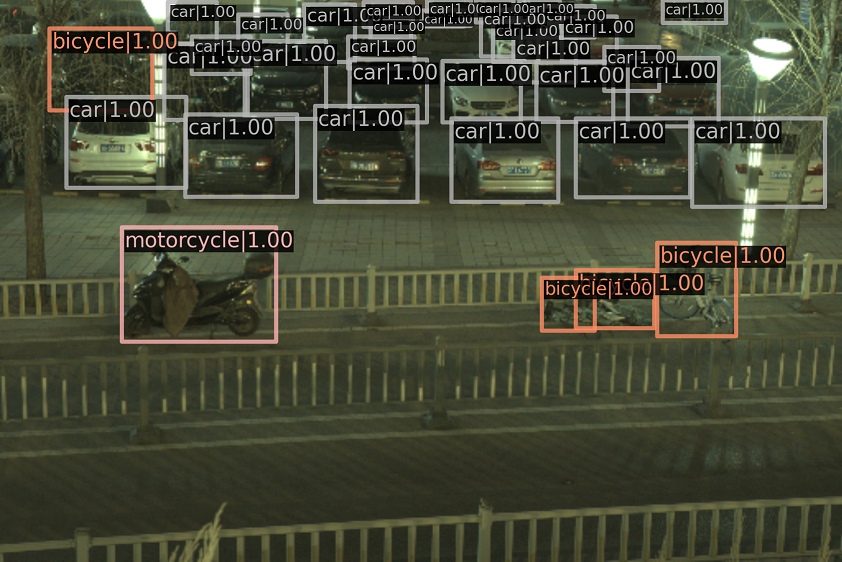}\tabularnewline
\scalebox{0.8}{diff. tuning (GM) \cite{yoshimura2022RAWAug}} & \scalebox{0.8}{NeuralAE (GM)\cite{onzon2021neural}} & \scalebox{0.8}{ours (AG+GM+CS)} & \scalebox{0.8}{GT}\tabularnewline
\end{tabular}

\vspace{4mm}

\caption{The visualization result on LODDataset \cite{Hong2021Crafting} trained
with simulated RAW-like data converted from COCO Dataset \cite{lin2014microsoft}.
The results of SID, Zero DCE, REDI, and H. Yang et. al. are from the
\cite{Hong2021Crafting}'s paper.}

\label{fig:viz_lod}
\end{figure*}

\begin{figure*}
\centering

\setlength{\tabcolsep}{0.5pt}


\begin{tabular}{cccc}
\includegraphics[width=0.245\textwidth]{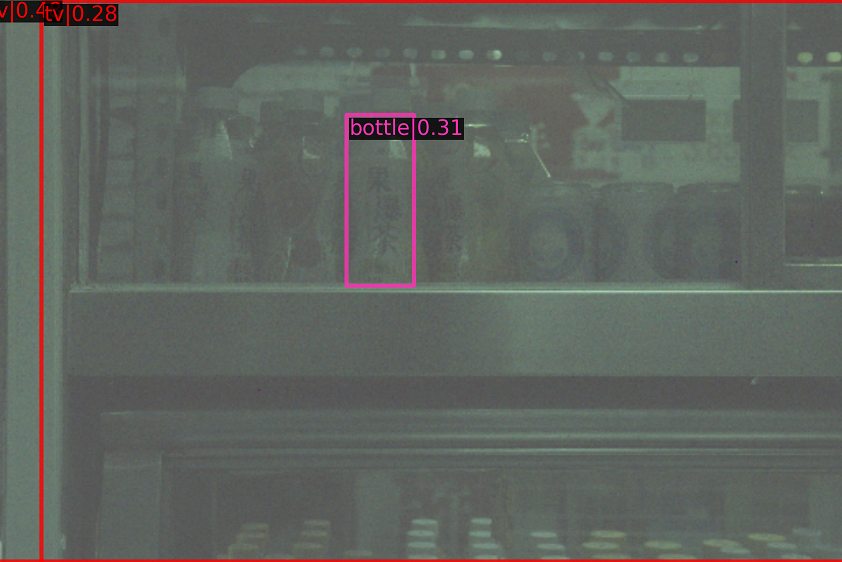} & \includegraphics[width=0.245\textwidth]{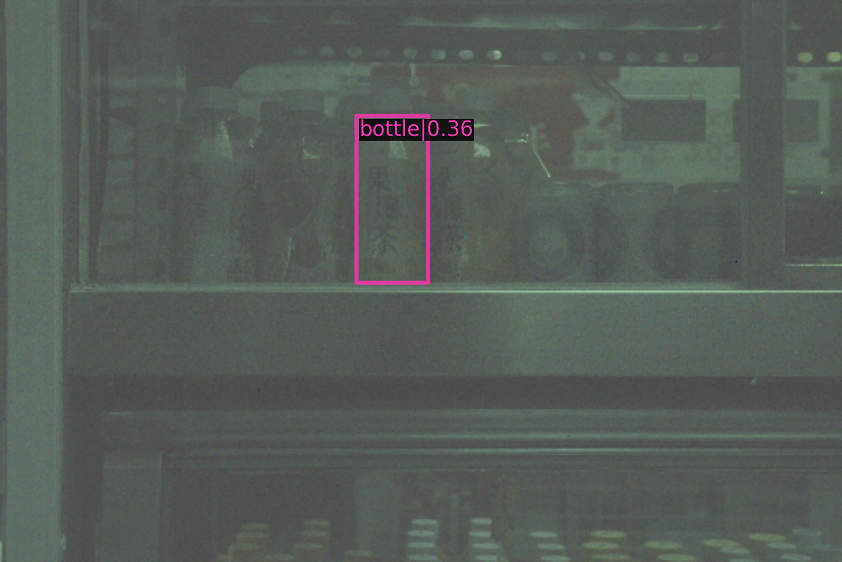} & \includegraphics[width=0.245\textwidth]{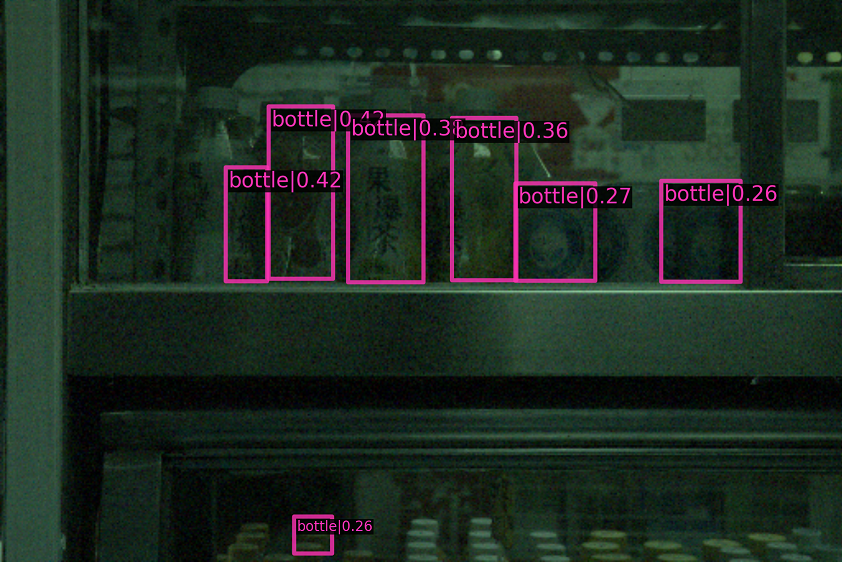} & \includegraphics[width=0.245\textwidth]{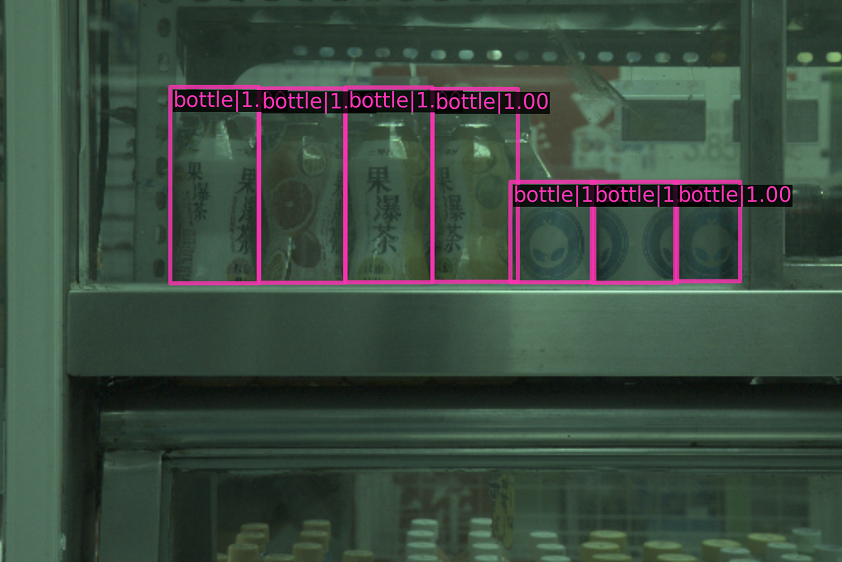}\tabularnewline
\scalebox{0.8}{diff. tuning (GM) \cite{yoshimura2022RAWAug}} & \scalebox{0.8}{NeuralAE (GM)\cite{onzon2021neural}} & \scalebox{0.8}{ours (AG+GM+CS)} & \scalebox{0.8}{GT}\tabularnewline
 &  &  & \tabularnewline
\includegraphics[width=0.245\textwidth]{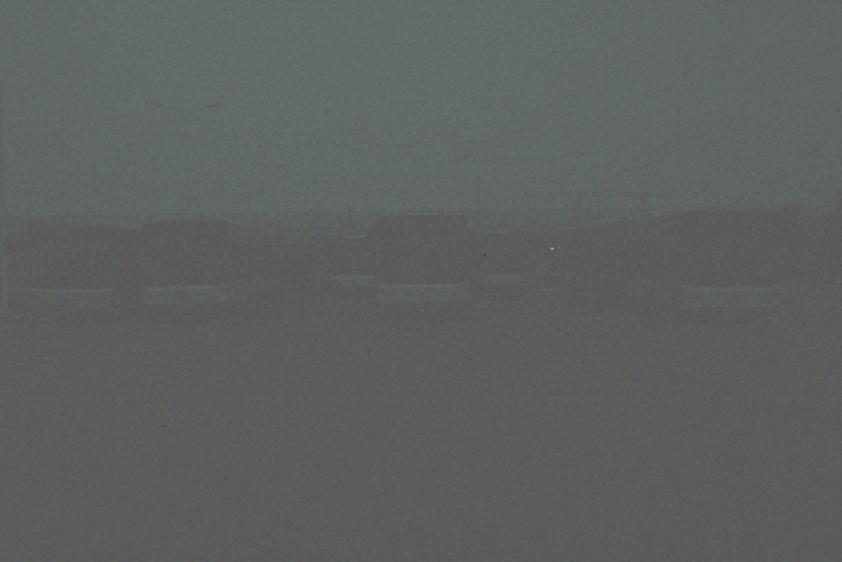} & \includegraphics[width=0.245\textwidth]{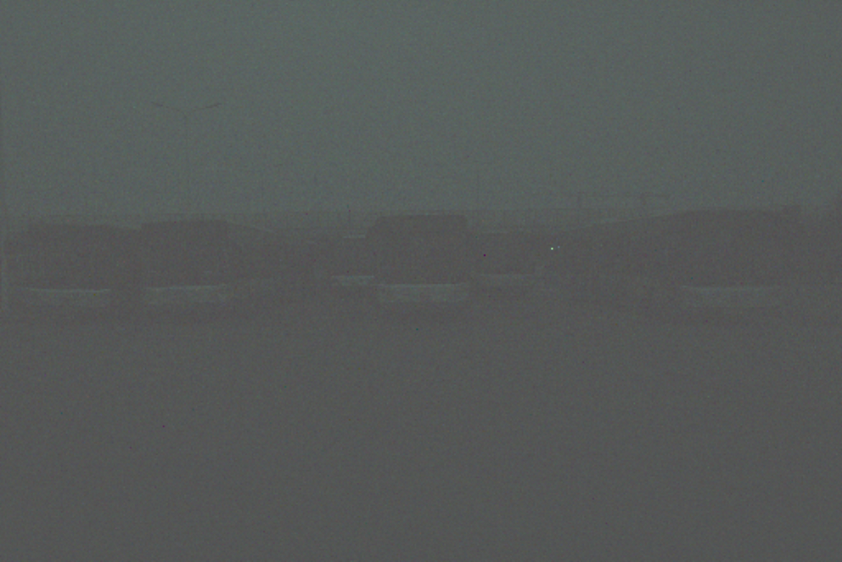} & \includegraphics[width=0.245\textwidth]{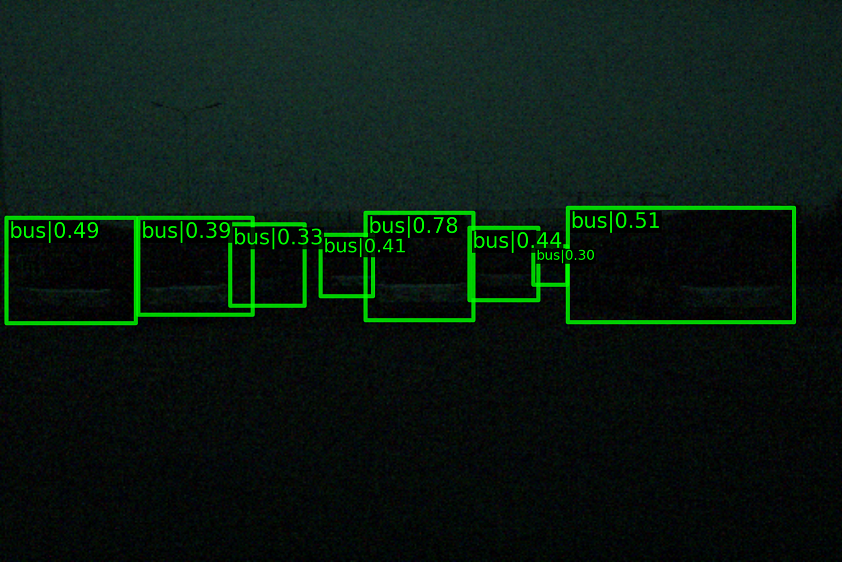} & \includegraphics[width=0.245\textwidth]{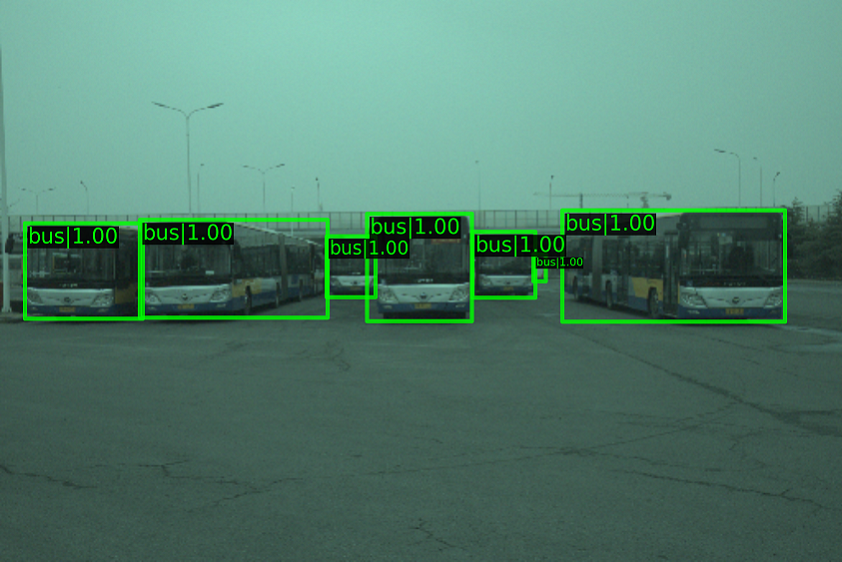}\tabularnewline
\scalebox{0.8}{diff. tuning (GM) \cite{yoshimura2022RAWAug}} & \scalebox{0.8}{NeuralAE (GM)\cite{onzon2021neural}} & \scalebox{0.8}{ours (AG+GM+CS)} & \scalebox{0.8}{GT}\tabularnewline
 &  &  & \tabularnewline
\includegraphics[width=0.245\textwidth]{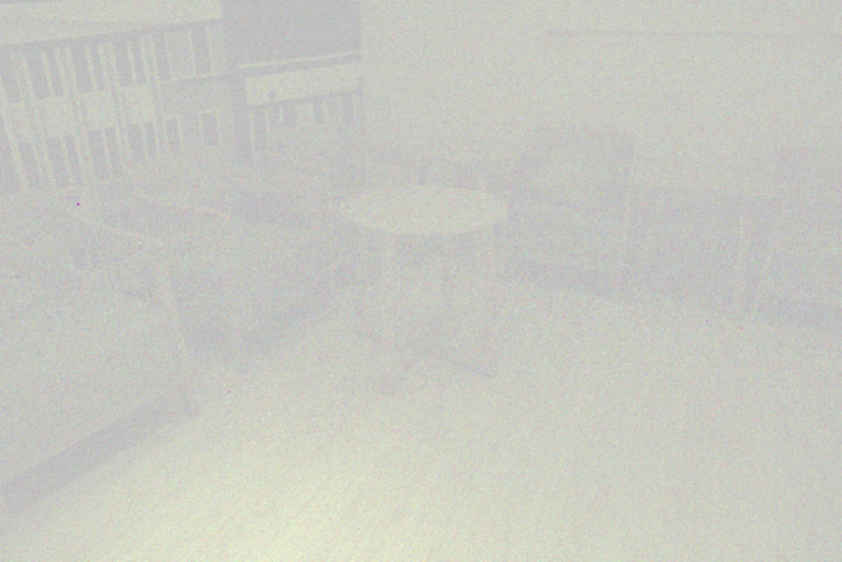} & \includegraphics[width=0.245\textwidth]{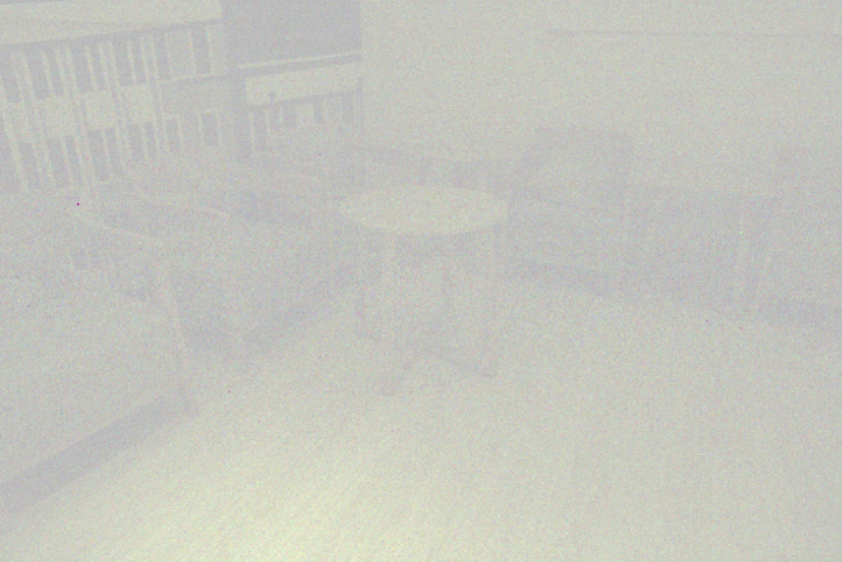} & \includegraphics[width=0.245\textwidth]{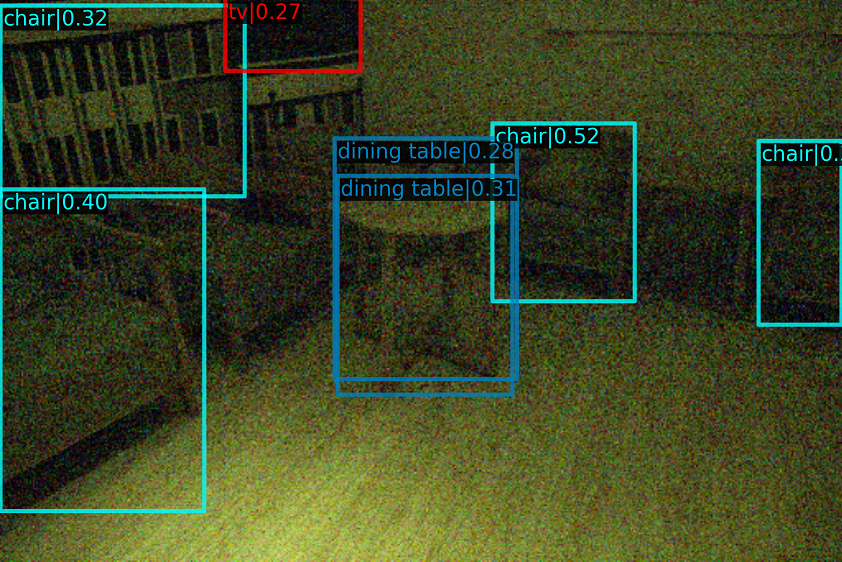} & \includegraphics[width=0.245\textwidth]{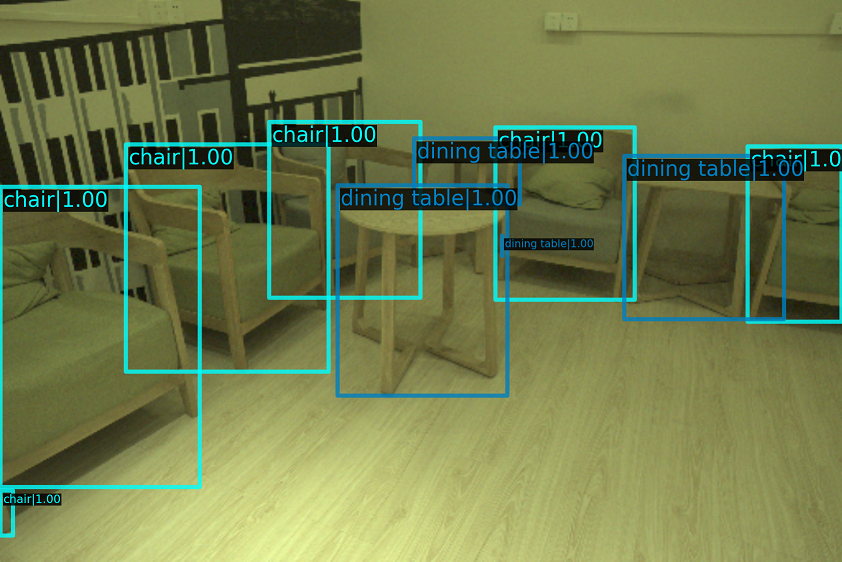}\tabularnewline
\scalebox{0.8}{diff. tuning (GM) \cite{yoshimura2022RAWAug}} & \scalebox{0.8}{NeuralAE (GM)\cite{onzon2021neural}} & \scalebox{0.8}{ours (AG+GM+CS)} & \scalebox{0.8}{GT}\tabularnewline
 &  &  & \tabularnewline
\includegraphics[width=0.245\textwidth]{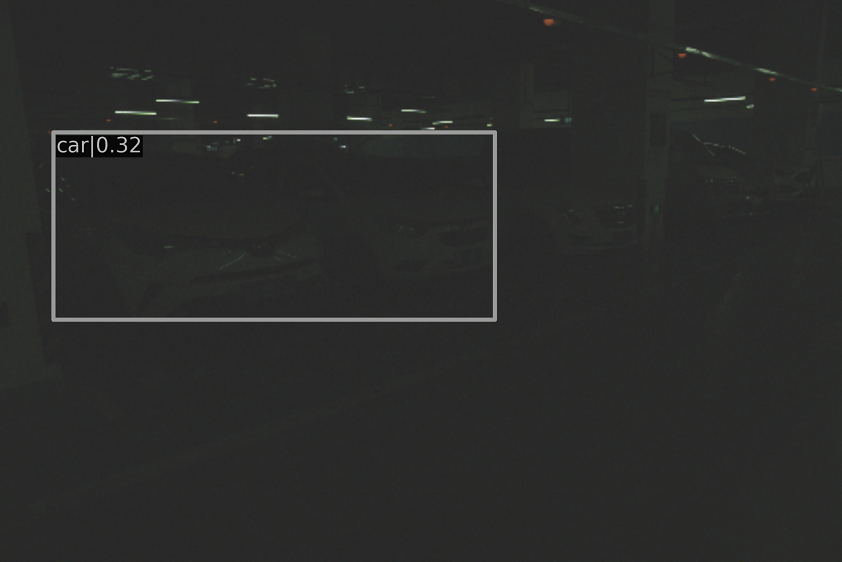} & \includegraphics[width=0.245\textwidth]{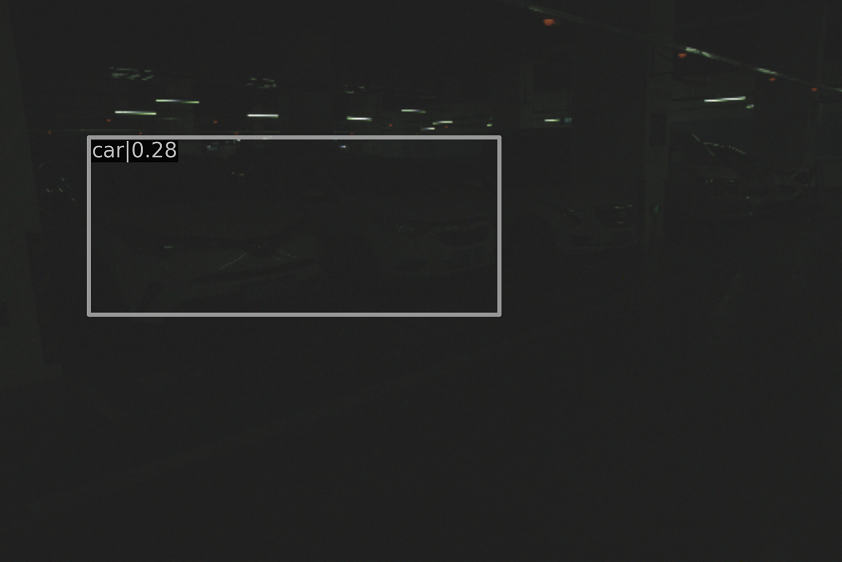} & \includegraphics[width=0.245\textwidth]{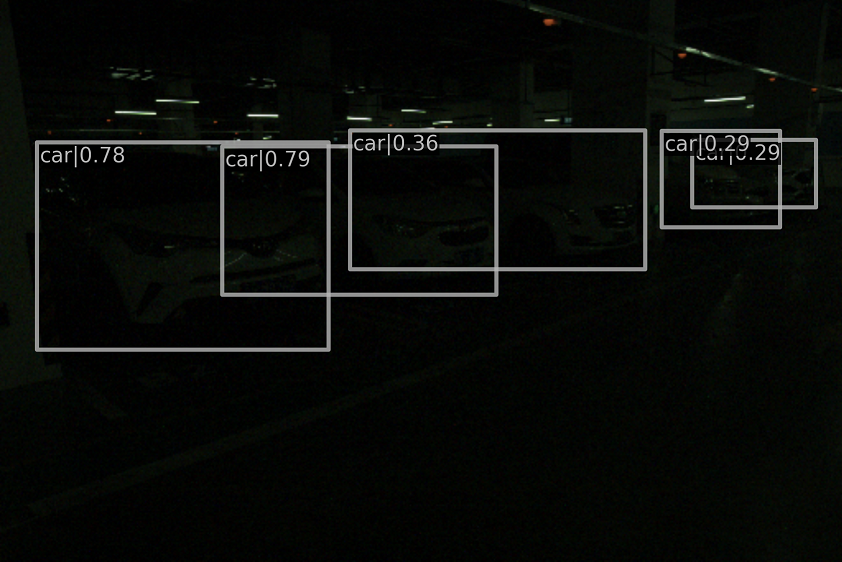} & \includegraphics[width=0.245\textwidth]{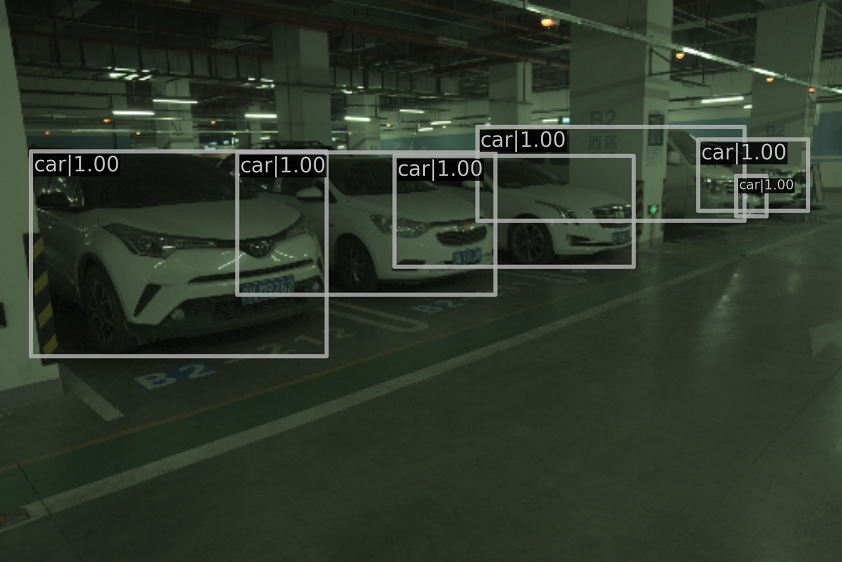}\tabularnewline
\scalebox{0.8}{diff. tuning (GM) \cite{yoshimura2022RAWAug}} & \scalebox{0.8}{NeuralAE (GM)\cite{onzon2021neural}} & \scalebox{0.8}{ours (AG+GM+CS)} & \scalebox{0.8}{GT}\tabularnewline
 &  &  & \tabularnewline
\includegraphics[width=0.245\textwidth]{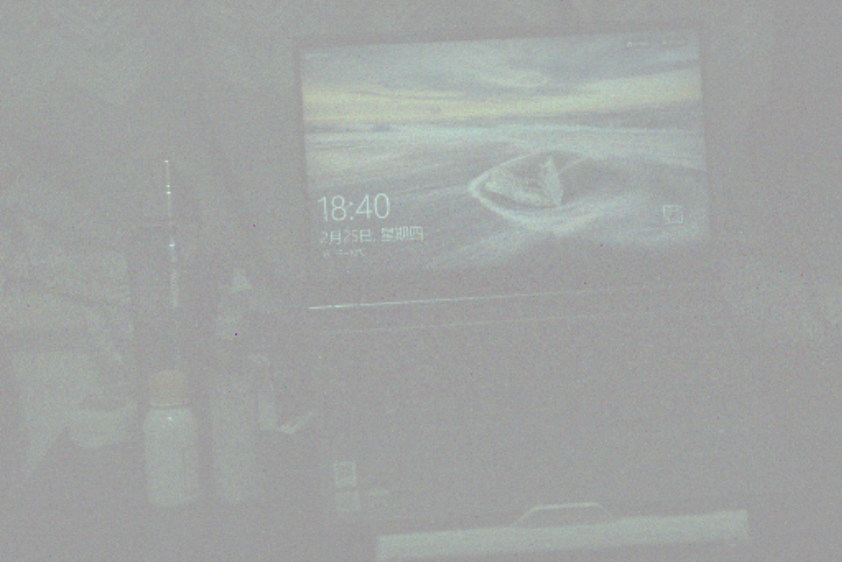} & \includegraphics[width=0.245\textwidth]{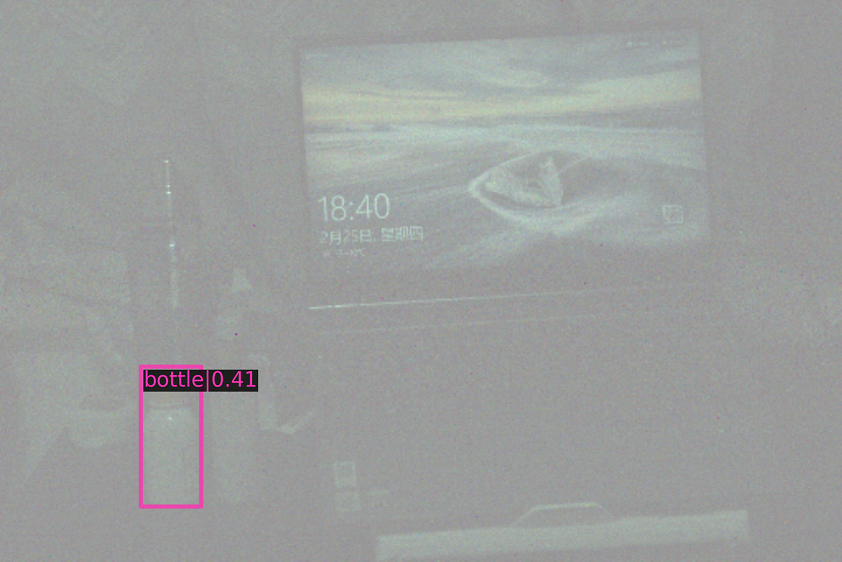} & \includegraphics[width=0.245\textwidth]{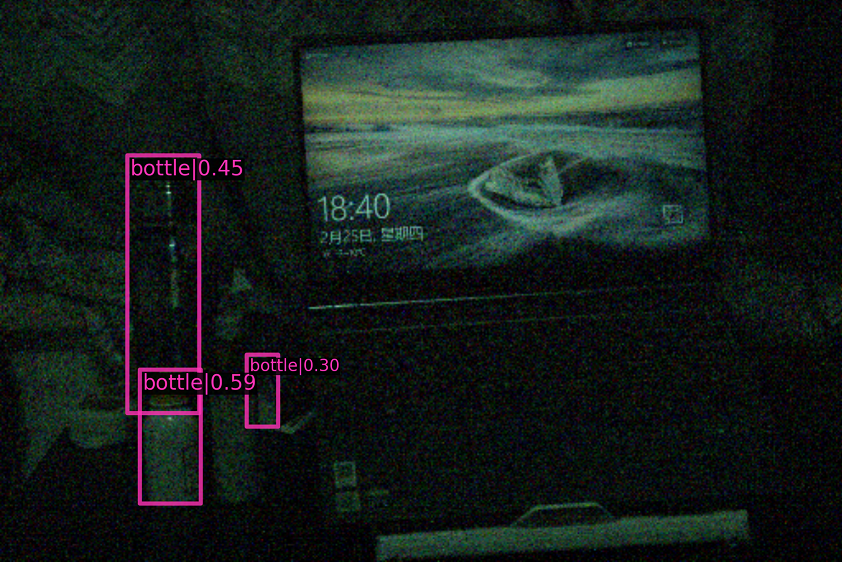} & \includegraphics[width=0.245\textwidth]{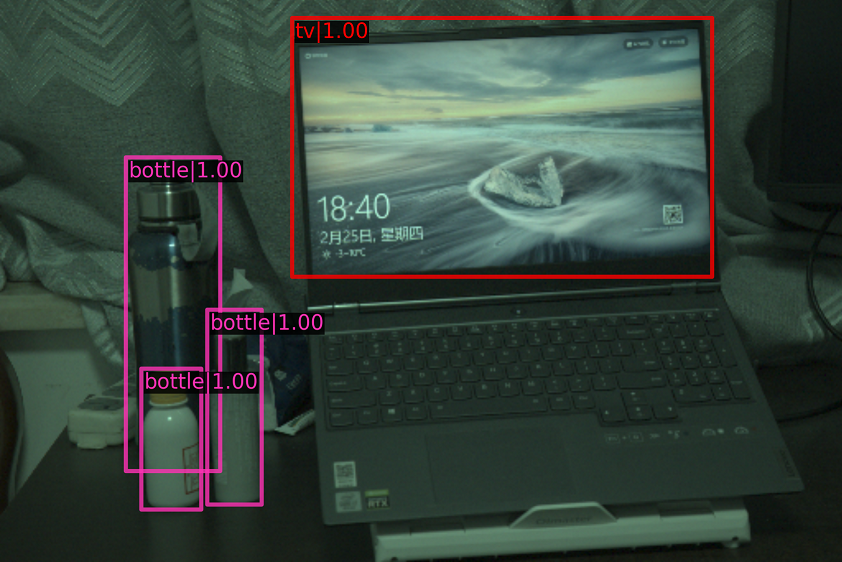}\tabularnewline
\scalebox{0.8}{diff. tuning (GM) \cite{yoshimura2022RAWAug}} & \scalebox{0.8}{NeuralAE (GM)\cite{onzon2021neural}} & \scalebox{0.8}{ours (AG+GM+CS)} & \scalebox{0.8}{GT}\tabularnewline
\end{tabular}

\vspace{4mm}

\caption{The visualization result on LODDataset \cite{Hong2021Crafting} trained
with simulated RAW-like data converted from COCO Dataset \cite{lin2014microsoft}.
More challenging images than images in Fig. \ref{fig:viz_lod} are
collected.}

\label{fig:viz_lod-1}
\end{figure*}

A more detailed comparison than Table 7 of the main paper is shown
in Table \ref{tab:coco2lod-1}. It is broken down by the level of
under-exposure. Our method obtains the highest accuracy among all
levels of under-exposure. The dynamic ISP control is able to convert
a broad luminance distribution environment to a preferable distribution
for the detector. The visualized comparison is in Fig. \ref{fig:viz_lod}
and Fig. \ref{fig:viz_lod-1}.

\subsubsection*{Training with Real Dark RAW Images}

We also evaluate the case of real RAW images used for training. The
real RAW images are randomly split into training data of 1830 images
and test data of 400, the same with \cite{Hong2021Crafting}. The
result is shown in Table \ref{tab:lod2lod}. Our method is confirmed
effective for small amounts of real RAW training data.

\begin{table}[h]
\caption{Evaluation on LODDataset \cite{Hong2021Crafting} trained with real
dark RAW data in LODDataset.}

\begin{center}

\begin{tabular}{cc|c}
\hline 
 & ISP & mAP@0.5:0.95\tabularnewline
\hline 
\hline 
H. Yang et. al. \cite{Hong2021Crafting} & - & 44.7\tabularnewline
NeuralAE \cite{onzon2021neural} & GM & 45.0\tabularnewline
NeuralAE \cite{onzon2021neural} & GM+CS & 45.5\tabularnewline
ours & GM & 45.4\tabularnewline
ours & AG+GM+CS & \textbf{46.2}\tabularnewline
\hline 
\end{tabular}

\end{center}

\label{tab:lod2lod}
\end{table}


\end{document}

%% file: figs/dynamicISP.pdf_tex
\begingroup%
  \makeatletter%
  \providecommand\color[2][]{%
    \errmessage{(Inkscape) Color is used for the text in Inkscape, but the package 'color.sty' is not loaded}%
    \renewcommand\color[2][]{}%
  }%
  \providecommand\transparent[1]{%
    \errmessage{(Inkscape) Transparency is used (non-zero) for the text in Inkscape, but the package 'transparent.sty' is not loaded}%
    \renewcommand\transparent[1]{}%
  }%
  \providecommand\rotatebox[2]{#2}%
  \newcommand*\fsize{\dimexpr\f@size pt\relax}%
  \newcommand*\lineheight[1]{\fontsize{\fsize}{#1\fsize}\selectfont}%
  \ifx\svgwidth\undefined%
    \setlength{\unitlength}{465.0637207bp}%
    \ifx\svgscale\undefined%
      \relax%
    \else%
      \setlength{\unitlength}{\unitlength * \real{\svgscale}}%
    \fi%
  \else%
    \setlength{\unitlength}{\svgwidth}%
  \fi%
  \global\let\svgwidth\undefined%
  \global\let\svgscale\undefined%
  \makeatother%
  \begin{picture}(1,0.36788314)%
    \lineheight{1}%
    \setlength\tabcolsep{0pt}%
    \put(0,0){\includegraphics[width=\unitlength,page=1]{dynamicISP.pdf}}%
    \put(0.11870893,0.2527601){\color[rgb]{1,1,1}\makebox(0,0)[lt]{\lineheight{1.25}\smash{\begin{tabular}[t]{l}Differentiable ISP\end{tabular}}}}%
    \put(0,0){\includegraphics[width=\unitlength,page=2]{dynamicISP.pdf}}%
    \put(0.53819219,0.25224173){\color[rgb]{1,1,1}\makebox(0,0)[lt]{\lineheight{1.25}\smash{\begin{tabular}[t]{l}DNN\end{tabular}}}}%
    \put(0,0){\includegraphics[width=\unitlength,page=3]{dynamicISP.pdf}}%
    \put(0.16929651,0.02285381){\color[rgb]{1,1,1}\makebox(0,0)[lt]{\lineheight{1.25}\smash{\begin{tabular}[t]{l}Controller \end{tabular}}}}%
    \put(0,0){\includegraphics[width=\unitlength,page=4]{dynamicISP.pdf}}%
    \put(0.1284849,0.15229843){\color[rgb]{0,0,0}\makebox(0,0)[lt]{\lineheight{1.25}\smash{\begin{tabular}[t]{l}Raise the tone \end{tabular}}}}%
    \put(0.13622577,0.1192707){\color[rgb]{0,0,0}\makebox(0,0)[lt]{\lineheight{1.25}\smash{\begin{tabular}[t]{l}on the stairs. \end{tabular}}}}%
    \put(0,0){\includegraphics[width=\unitlength,page=5]{dynamicISP.pdf}}%
    \put(0.49527336,0.09742421){\color[rgb]{0,0,0}\makebox(0,0)[lt]{\lineheight{1.25}\smash{\begin{tabular}[t]{l}The stairs are\end{tabular}}}}%
    \put(0.49527334,0.0643965){\color[rgb]{0,0,0}\makebox(0,0)[lt]{\lineheight{1.25}\smash{\begin{tabular}[t]{l}too dark to see.\end{tabular}}}}%
  \end{picture}%
\endgroup%

%% file: figs/lattent_update_style.pdf_tex
\begingroup%
  \makeatletter%
  \providecommand\color[2][]{%
    \errmessage{(Inkscape) Color is used for the text in Inkscape, but the package 'color.sty' is not loaded}%
    \renewcommand\color[2][]{}%
  }%
  \providecommand\transparent[1]{%
    \errmessage{(Inkscape) Transparency is used (non-zero) for the text in Inkscape, but the package 'transparent.sty' is not loaded}%
    \renewcommand\transparent[1]{}%
  }%
  \providecommand\rotatebox[2]{#2}%
  \newcommand*\fsize{\dimexpr\f@size pt\relax}%
  \newcommand*\lineheight[1]{\fontsize{\fsize}{#1\fsize}\selectfont}%
  \ifx\svgwidth\undefined%
    \setlength{\unitlength}{960bp}%
    \ifx\svgscale\undefined%
      \relax%
    \else%
      \setlength{\unitlength}{\unitlength * \real{\svgscale}}%
    \fi%
  \else%
    \setlength{\unitlength}{\svgwidth}%
  \fi%
  \global\let\svgwidth\undefined%
  \global\let\svgscale\undefined%
  \makeatother%
  \begin{picture}(1,0.36209335)%
    \lineheight{1}%
    \setlength\tabcolsep{0pt}%
    \put(0,0){\includegraphics[width=\unitlength,page=1]{lattent_update_style.pdf}}%
    \put(0.33393749,0.3424724){\color[rgb]{0,0,0}\makebox(0,0)[lt]{\lineheight{1.25}\smash{\begin{tabular}[t]{l}Differentiable ISP\end{tabular}}}}%
    \put(0,0){\includegraphics[width=\unitlength,page=2]{lattent_update_style.pdf}}%
    \put(0.6960104,0.2525349){\color[rgb]{0,0,0}\makebox(0,0)[lt]{\lineheight{1.25}\smash{\begin{tabular}[t]{l}Detector\end{tabular}}}}%
    \put(0,0){\includegraphics[width=\unitlength,page=3]{lattent_update_style.pdf}}%
    \put(0.6996354,0.08386824){\color[rgb]{0,0,0}\makebox(0,0)[lt]{\lineheight{1.25}\smash{\begin{tabular}[t]{l}$f_{SFB}$\end{tabular}}}}%
    \put(0,0){\includegraphics[width=\unitlength,page=4]{lattent_update_style.pdf}}%
    \put(0.40756137,0.29891796){\color[rgb]{0,0,0}\rotatebox{-90.00000252}{\makebox(0,0)[lt]{\lineheight{1.25}\smash{\begin{tabular}[t]{l}Function 2\end{tabular}}}}}%
    \put(0,0){\includegraphics[width=\unitlength,page=5]{lattent_update_style.pdf}}%
    \put(0.58999999,0.30011823){\color[rgb]{0,0,0}\rotatebox{-90.00000252}{\makebox(0,0)[lt]{\lineheight{1.25}\smash{\begin{tabular}[t]{l}Function  N\end{tabular}}}}}%
    \put(0,0){\includegraphics[width=\unitlength,page=6]{lattent_update_style.pdf}}%
    \put(0.21126041,0.11602449){\color[rgb]{0,0,0}\makebox(0,0)[lt]{\lineheight{1.25}\smash{\begin{tabular}[t]{l}$f_{de,1}$\end{tabular}}}}%
    \put(0,0){\includegraphics[width=\unitlength,page=7]{lattent_update_style.pdf}}%
    \put(0.38633332,0.11527449){\color[rgb]{0,0,0}\makebox(0,0)[lt]{\lineheight{1.25}\smash{\begin{tabular}[t]{l}$f_{de,2}$\end{tabular}}}}%
    \put(0,0){\includegraphics[width=\unitlength,page=8]{lattent_update_style.pdf}}%
    \put(0.56961457,0.1159724){\color[rgb]{0,0,0}\makebox(0,0)[lt]{\lineheight{1.25}\smash{\begin{tabular}[t]{l}$f_{de,N}$\end{tabular}}}}%
    \put(0,0){\includegraphics[width=\unitlength,page=9]{lattent_update_style.pdf}}%
    \put(0.1906875,0.15707657){\color[rgb]{0,0,0}\makebox(0,0)[lt]{\lineheight{1.25}\smash{\begin{tabular}[t]{l}decide param.\end{tabular}}}}%
    \put(0,0){\includegraphics[width=\unitlength,page=10]{lattent_update_style.pdf}}%
    \put(0.27904166,0.11602449){\color[rgb]{0,0,0}\makebox(0,0)[lt]{\lineheight{1.25}\smash{\begin{tabular}[t]{l}$f_{up,1}$\end{tabular}}}}%
    \put(0,0){\includegraphics[width=\unitlength,page=11]{lattent_update_style.pdf}}%
    \put(0.45295832,0.11561824){\color[rgb]{0,0,0}\makebox(0,0)[lt]{\lineheight{1.25}\smash{\begin{tabular}[t]{l}$f_{up,2}$\end{tabular}}}}%
    \put(0,0){\includegraphics[width=\unitlength,page=12]{lattent_update_style.pdf}}%
    \put(0.87342706,0.0114724){\color[rgb]{0,0,0}\makebox(0,0)[lt]{\lineheight{1.25}\smash{\begin{tabular}[t]{l}: latent variable\end{tabular}}}}%
    \put(0,0){\includegraphics[width=\unitlength,page=13]{lattent_update_style.pdf}}%
    \put(0.23144791,0.29862743){\color[rgb]{0,0,0}\rotatebox{-90.00000252}{\makebox(0,0)[lt]{\lineheight{1.25}\smash{\begin{tabular}[t]{l}Function 1\end{tabular}}}}}%
    \put(0,0){\includegraphics[width=\unitlength,page=14]{lattent_update_style.pdf}}%
    \put(0.45701041,0.01277449){\color[rgb]{0,0,0}\makebox(0,0)[lt]{\lineheight{1.25}\smash{\begin{tabular}[t]{l}Controller\end{tabular}}}}%
    \put(0.30870833,0.21329532){\color[rgb]{0,0,0}\makebox(0,0)[lt]{\lineheight{1.25}\smash{\begin{tabular}[t]{l}update \end{tabular}}}}%
    \put(0.30733333,0.19579532){\color[rgb]{0,0,0}\makebox(0,0)[lt]{\lineheight{1.25}\smash{\begin{tabular}[t]{l}latent param.\end{tabular}}}}%
    \put(0,0){\includegraphics[width=\unitlength,page=15]{lattent_update_style.pdf}}%
    \put(0.08050742,0.17892504){\color[rgb]{0,0,0}\makebox(0,0)[t]{\lineheight{1.25}\smash{\begin{tabular}[t]{c}RAW image\end{tabular}}}}%
  \end{picture}%
\endgroup%

%% file: figs/training.pdf_tex
\begingroup%
  \makeatletter%
  \providecommand\color[2][]{%
    \errmessage{(Inkscape) Color is used for the text in Inkscape, but the package 'color.sty' is not loaded}%
    \renewcommand\color[2][]{}%
  }%
  \providecommand\transparent[1]{%
    \errmessage{(Inkscape) Transparency is used (non-zero) for the text in Inkscape, but the package 'transparent.sty' is not loaded}%
    \renewcommand\transparent[1]{}%
  }%
  \providecommand\rotatebox[2]{#2}%
  \newcommand*\fsize{\dimexpr\f@size pt\relax}%
  \newcommand*\lineheight[1]{\fontsize{\fsize}{#1\fsize}\selectfont}%
  \ifx\svgwidth\undefined%
    \setlength{\unitlength}{642.26303101bp}%
    \ifx\svgscale\undefined%
      \relax%
    \else%
      \setlength{\unitlength}{\unitlength * \real{\svgscale}}%
    \fi%
  \else%
    \setlength{\unitlength}{\svgwidth}%
  \fi%
  \global\let\svgwidth\undefined%
  \global\let\svgscale\undefined%
  \makeatother%
  \begin{picture}(1,0.39323471)%
    \lineheight{1}%
    \setlength\tabcolsep{0pt}%
    \put(0,0){\includegraphics[width=\unitlength,page=1]{training.pdf}}%
    \put(0.31203696,0.10863024){\color[rgb]{1,1,1}\makebox(0,0)[lt]{\lineheight{1.25}\smash{\begin{tabular}[t]{l}Differentiable ISP\end{tabular}}}}%
    \put(0,0){\includegraphics[width=\unitlength,page=2]{training.pdf}}%
    \put(0.63097171,0.10863024){\color[rgb]{1,1,1}\makebox(0,0)[lt]{\lineheight{1.25}\smash{\begin{tabular}[t]{l}DNN\end{tabular}}}}%
    \put(0,0){\includegraphics[width=\unitlength,page=3]{training.pdf}}%
    \put(0.82354078,0.03489098){\color[rgb]{0,0,0}\makebox(0,0)[lt]{\lineheight{1.25}\smash{\begin{tabular}[t]{l}Easy to detect\end{tabular}}}}%
    \put(0,0){\includegraphics[width=\unitlength,page=4]{training.pdf}}%
    \put(0.63950404,0.03159016){\color[rgb]{0,0,0}\makebox(0,0)[lt]{\lineheight{1.25}\smash{\begin{tabular}[t]{l}Detection loss\end{tabular}}}}%
    \put(0,0){\includegraphics[width=\unitlength,page=5]{training.pdf}}%
    \put(0.18719715,0.04616362){\color[rgb]{0,0,0}\makebox(0,0)[lt]{\lineheight{1.25}\smash{\begin{tabular}[t]{l}Backpropagate through all the \end{tabular}}}}%
    \put(0.29724552,0.01626933){\color[rgb]{0,0,0}\makebox(0,0)[lt]{\lineheight{1.25}\smash{\begin{tabular}[t]{l}path above \end{tabular}}}}%
    \put(0.79629342,0.19987009){\color[rgb]{0,0,0}\makebox(0,0)[lt]{\lineheight{1.25}\smash{\begin{tabular}[t]{l}Difficult to detect\end{tabular}}}}%
    \put(0,0){\includegraphics[width=\unitlength,page=6]{training.pdf}}%
    \put(0.01904175,0.2170905){\color[rgb]{0,0,0}\makebox(0,0)[lt]{\lineheight{1.25}\smash{\begin{tabular}[t]{l}First frame\end{tabular}}}}%
    \put(0.00256875,0.04550969){\color[rgb]{0,0,0}\makebox(0,0)[lt]{\lineheight{1.25}\smash{\begin{tabular}[t]{l}Second frame\end{tabular}}}}%
    \put(0,0){\includegraphics[width=\unitlength,page=7]{training.pdf}}%
    \put(0.30054634,0.36126816){\color[rgb]{1,1,1}\makebox(0,0)[lt]{\lineheight{1.25}\smash{\begin{tabular}[t]{l}Parameter Initializer\end{tabular}}}}%
    \put(0,0){\includegraphics[width=\unitlength,page=8]{training.pdf}}%
    \put(0.31203696,0.27908999){\color[rgb]{1,1,1}\makebox(0,0)[lt]{\lineheight{1.25}\smash{\begin{tabular}[t]{l}Differentiable ISP\end{tabular}}}}%
    \put(0,0){\includegraphics[width=\unitlength,page=9]{training.pdf}}%
    \put(0.35616219,0.19968328){\color[rgb]{1,1,1}\makebox(0,0)[lt]{\lineheight{1.25}\smash{\begin{tabular}[t]{l}Controller\end{tabular}}}}%
    \put(0,0){\includegraphics[width=\unitlength,page=10]{training.pdf}}%
    \put(0.63078487,0.27843606){\color[rgb]{1,1,1}\makebox(0,0)[lt]{\lineheight{1.25}\smash{\begin{tabular}[t]{l}DNN\end{tabular}}}}%
  \end{picture}%
\endgroup%

%% file: figs/controller_comparison2.pdf_tex
\begingroup%
  \makeatletter%
  \providecommand\color[2][]{%
    \errmessage{(Inkscape) Color is used for the text in Inkscape, but the package 'color.sty' is not loaded}%
    \renewcommand\color[2][]{}%
  }%
  \providecommand\transparent[1]{%
    \errmessage{(Inkscape) Transparency is used (non-zero) for the text in Inkscape, but the package 'transparent.sty' is not loaded}%
    \renewcommand\transparent[1]{}%
  }%
  \providecommand\rotatebox[2]{#2}%
  \newcommand*\fsize{\dimexpr\f@size pt\relax}%
  \newcommand*\lineheight[1]{\fontsize{\fsize}{#1\fsize}\selectfont}%
  \ifx\svgwidth\undefined%
    \setlength{\unitlength}{709.56719971bp}%
    \ifx\svgscale\undefined%
      \relax%
    \else%
      \setlength{\unitlength}{\unitlength * \real{\svgscale}}%
    \fi%
  \else%
    \setlength{\unitlength}{\svgwidth}%
  \fi%
  \global\let\svgwidth\undefined%
  \global\let\svgscale\undefined%
  \makeatother%
  \begin{picture}(1,0.49105916)%
    \lineheight{1}%
    \setlength\tabcolsep{0pt}%
    \put(0,0){\includegraphics[width=\unitlength,page=1]{controller_comparison2.pdf}}%
    \put(0.42079267,0.46044582){\color[rgb]{0,0,0}\makebox(0,0)[lt]{\lineheight{1.25}\smash{\begin{tabular}[t]{l}Differentiable ISP\end{tabular}}}}%
    \put(0,0){\includegraphics[width=\unitlength,page=2]{controller_comparison2.pdf}}%
    \put(0.85356351,0.34116184){\color[rgb]{0,0,0}\makebox(0,0)[lt]{\lineheight{1.25}\smash{\begin{tabular}[t]{l}Detector\end{tabular}}}}%
    \put(0,0){\includegraphics[width=\unitlength,page=3]{controller_comparison2.pdf}}%
    \put(0.528929,0.4073994){\color[rgb]{0,0,0}\rotatebox{-90.00000252}{\makebox(0,0)[lt]{\lineheight{1.25}\smash{\begin{tabular}[t]{l}F\end{tabular}}}}}%
    \put(0.528929,0.39184062){\color[rgb]{0,0,0}\rotatebox{-90.00000252}{\makebox(0,0)[lt]{\lineheight{1.25}\smash{\begin{tabular}[t]{l}unction 2\end{tabular}}}}}%
    \put(0,0){\includegraphics[width=\unitlength,page=4]{controller_comparison2.pdf}}%
    \put(0.76663728,0.40948518){\color[rgb]{0,0,0}\rotatebox{-90.00000252}{\makebox(0,0)[lt]{\lineheight{1.25}\smash{\begin{tabular}[t]{l}Function N\end{tabular}}}}}%
    \put(0,0){\includegraphics[width=\unitlength,page=5]{controller_comparison2.pdf}}%
    \put(0.28520298,0.15618994){\color[rgb]{1,1,1}\makebox(0,0)[lt]{\lineheight{1.25}\smash{\begin{tabular}[t]{l}DNN\end{tabular}}}}%
    \put(0,0){\includegraphics[width=\unitlength,page=6]{controller_comparison2.pdf}}%
    \put(0.30132548,0.40796313){\color[rgb]{0,0,0}\rotatebox{-90.00000252}{\makebox(0,0)[lt]{\lineheight{1.25}\smash{\begin{tabular}[t]{l}F\end{tabular}}}}}%
    \put(0.30132548,0.39240435){\color[rgb]{0,0,0}\rotatebox{-90.00000252}{\makebox(0,0)[lt]{\lineheight{1.25}\smash{\begin{tabular}[t]{l}unction 1\end{tabular}}}}}%
    \put(0,0){\includegraphics[width=\unitlength,page=7]{controller_comparison2.pdf}}%
    \put(0.45570127,0.01479389){\color[rgb]{0,0,0}\makebox(0,0)[lt]{\lineheight{1.25}\smash{\begin{tabular}[t]{l}Controller\end{tabular}}}}%
    \put(0,0){\includegraphics[width=\unitlength,page=8]{controller_comparison2.pdf}}%
    \put(0.28280715,0.15414644){\color[rgb]{0,0,0}\makebox(0,0)[lt]{\lineheight{1.25}\smash{\begin{tabular}[t]{l}FCs\end{tabular}}}}%
    \put(0,0){\includegraphics[width=\unitlength,page=9]{controller_comparison2.pdf}}%
    \put(0.51084756,0.15434374){\color[rgb]{0,0,0}\makebox(0,0)[lt]{\lineheight{1.25}\smash{\begin{tabular}[t]{l}FCs\end{tabular}}}}%
    \put(0,0){\includegraphics[width=\unitlength,page=10]{controller_comparison2.pdf}}%
    \put(0.74811895,0.15255392){\color[rgb]{0,0,0}\makebox(0,0)[lt]{\lineheight{1.25}\smash{\begin{tabular}[t]{l}FCs\end{tabular}}}}%
    \put(0.19176574,0.07784641){\color[rgb]{0,0,0}\makebox(0,0)[lt]{\lineheight{1.25}\smash{\begin{tabular}[t]{l}Encoder\end{tabular}}}}%
    \put(0,0){\includegraphics[width=\unitlength,page=11]{controller_comparison2.pdf}}%
  \end{picture}%
\endgroup%

%% file: figs/controller_comparison3.pdf_tex
\begingroup%
  \makeatletter%
  \providecommand\color[2][]{%
    \errmessage{(Inkscape) Color is used for the text in Inkscape, but the package 'color.sty' is not loaded}%
    \renewcommand\color[2][]{}%
  }%
  \providecommand\transparent[1]{%
    \errmessage{(Inkscape) Transparency is used (non-zero) for the text in Inkscape, but the package 'transparent.sty' is not loaded}%
    \renewcommand\transparent[1]{}%
  }%
  \providecommand\rotatebox[2]{#2}%
  \newcommand*\fsize{\dimexpr\f@size pt\relax}%
  \newcommand*\lineheight[1]{\fontsize{\fsize}{#1\fsize}\selectfont}%
  \ifx\svgwidth\undefined%
    \setlength{\unitlength}{726.03881836bp}%
    \ifx\svgscale\undefined%
      \relax%
    \else%
      \setlength{\unitlength}{\unitlength * \real{\svgscale}}%
    \fi%
  \else%
    \setlength{\unitlength}{\svgwidth}%
  \fi%
  \global\let\svgwidth\undefined%
  \global\let\svgscale\undefined%
  \makeatother%
  \begin{picture}(1,0.4809886)%
    \lineheight{1}%
    \setlength\tabcolsep{0pt}%
    \put(0,0){\includegraphics[width=\unitlength,page=1]{controller_comparison3.pdf}}%
    \put(0.8700085,0.12431813){\color[rgb]{0,0,0}\makebox(0,0)[lt]{\lineheight{1.25}\smash{\begin{tabular}[t]{l}1D tensor\\operations\end{tabular}}}}%
    \put(0.87065532,0.04634695){\color[rgb]{0,0,0}\makebox(0,0)[lt]{\lineheight{1.25}\smash{\begin{tabular}[t]{l}3D tensor\\operations\end{tabular}}}}%
    \put(0,0){\includegraphics[width=\unitlength,page=2]{controller_comparison3.pdf}}%
    \put(0.40918016,0.45011924){\color[rgb]{0,0,0}\makebox(0,0)[lt]{\lineheight{1.25}\smash{\begin{tabular}[t]{l}Differentiable ISP\end{tabular}}}}%
    \put(0,0){\includegraphics[width=\unitlength,page=3]{controller_comparison3.pdf}}%
    \put(0.83815171,0.33354145){\color[rgb]{0,0,0}\makebox(0,0)[lt]{\lineheight{1.25}\smash{\begin{tabular}[t]{l}Detector\end{tabular}}}}%
    \put(0,0){\includegraphics[width=\unitlength,page=4]{controller_comparison3.pdf}}%
    \put(0.51692922,0.39827628){\color[rgb]{0,0,0}\rotatebox{-90.00000252}{\makebox(0,0)[lt]{\lineheight{1.25}\smash{\begin{tabular}[t]{l}F\end{tabular}}}}}%
    \put(0.51692922,0.38307049){\color[rgb]{0,0,0}\rotatebox{-90.00000252}{\makebox(0,0)[lt]{\lineheight{1.25}\smash{\begin{tabular}[t]{l}unction 2\end{tabular}}}}}%
    \put(0,0){\includegraphics[width=\unitlength,page=5]{controller_comparison3.pdf}}%
    \put(0.74924462,0.40031474){\color[rgb]{0,0,0}\rotatebox{-90.00000252}{\makebox(0,0)[lt]{\lineheight{1.25}\smash{\begin{tabular}[t]{l}Function N\end{tabular}}}}}%
    \put(0,0){\includegraphics[width=\unitlength,page=6]{controller_comparison3.pdf}}%
    \put(0.27873259,0.152766){\color[rgb]{1,1,1}\makebox(0,0)[lt]{\lineheight{1.25}\smash{\begin{tabular}[t]{l}DNN\end{tabular}}}}%
    \put(0,0){\includegraphics[width=\unitlength,page=7]{controller_comparison3.pdf}}%
    \put(0.29448933,0.39882722){\color[rgb]{0,0,0}\rotatebox{-90.00000252}{\makebox(0,0)[lt]{\lineheight{1.25}\smash{\begin{tabular}[t]{l}F\end{tabular}}}}}%
    \put(0.29448933,0.38362142){\color[rgb]{0,0,0}\rotatebox{-90.00000252}{\makebox(0,0)[lt]{\lineheight{1.25}\smash{\begin{tabular}[t]{l}unction 1\end{tabular}}}}}%
    \put(0,0){\includegraphics[width=\unitlength,page=8]{controller_comparison3.pdf}}%
    \put(0.45200156,0.0145778){\color[rgb]{0,0,0}\makebox(0,0)[lt]{\lineheight{1.25}\smash{\begin{tabular}[t]{l}Controller\end{tabular}}}}%
    \put(0,0){\includegraphics[width=\unitlength,page=9]{controller_comparison3.pdf}}%
    \put(0.27639112,0.14870286){\color[rgb]{0,0,0}\makebox(0,0)[lt]{\lineheight{1.25}\smash{\begin{tabular}[t]{l}FCs\end{tabular}}}}%
    \put(0,0){\includegraphics[width=\unitlength,page=10]{controller_comparison3.pdf}}%
    \put(0.49925799,0.14889568){\color[rgb]{0,0,0}\makebox(0,0)[lt]{\lineheight{1.25}\smash{\begin{tabular}[t]{l}FCs\end{tabular}}}}%
    \put(0,0){\includegraphics[width=\unitlength,page=11]{controller_comparison3.pdf}}%
    \put(0.73114641,0.14714647){\color[rgb]{0,0,0}\makebox(0,0)[lt]{\lineheight{1.25}\smash{\begin{tabular}[t]{l}FCs\end{tabular}}}}%
    \put(0.18741517,0.07619985){\color[rgb]{0,0,0}\makebox(0,0)[lt]{\lineheight{1.25}\smash{\begin{tabular}[t]{l}Encoder\end{tabular}}}}%
    \put(0,0){\includegraphics[width=\unitlength,page=12]{controller_comparison3.pdf}}%
    \put(0.41030958,0.07442308){\color[rgb]{0,0,0}\makebox(0,0)[lt]{\lineheight{1.25}\smash{\begin{tabular}[t]{l}Encoder\end{tabular}}}}%
    \put(0,0){\includegraphics[width=\unitlength,page=13]{controller_comparison3.pdf}}%
    \put(0.64347892,0.07697116){\color[rgb]{0,0,0}\makebox(0,0)[lt]{\lineheight{1.25}\smash{\begin{tabular}[t]{l}Encoder\end{tabular}}}}%
    \put(0,0){\includegraphics[width=\unitlength,page=14]{controller_comparison3.pdf}}%
  \end{picture}%
\endgroup%

%% file: figs/controller_comparison4.pdf_tex
\begingroup%
  \makeatletter%
  \providecommand\color[2][]{%
    \errmessage{(Inkscape) Color is used for the text in Inkscape, but the package 'color.sty' is not loaded}%
    \renewcommand\color[2][]{}%
  }%
  \providecommand\transparent[1]{%
    \errmessage{(Inkscape) Transparency is used (non-zero) for the text in Inkscape, but the package 'transparent.sty' is not loaded}%
    \renewcommand\transparent[1]{}%
  }%
  \providecommand\rotatebox[2]{#2}%
  \newcommand*\fsize{\dimexpr\f@size pt\relax}%
  \newcommand*\lineheight[1]{\fontsize{\fsize}{#1\fsize}\selectfont}%
  \ifx\svgwidth\undefined%
    \setlength{\unitlength}{708.78501892bp}%
    \ifx\svgscale\undefined%
      \relax%
    \else%
      \setlength{\unitlength}{\unitlength * \real{\svgscale}}%
    \fi%
  \else%
    \setlength{\unitlength}{\svgwidth}%
  \fi%
  \global\let\svgwidth\undefined%
  \global\let\svgscale\undefined%
  \makeatother%
  \begin{picture}(1,0.53748785)%
    \lineheight{1}%
    \setlength\tabcolsep{0pt}%
    \put(0,0){\includegraphics[width=\unitlength,page=1]{controller_comparison4.pdf}}%
    \put(0.41614368,0.50623475){\color[rgb]{0,0,0}\makebox(0,0)[lt]{\lineheight{1.25}\smash{\begin{tabular}[t]{l}Differentiable ISP\end{tabular}}}}%
    \put(0,0){\includegraphics[width=\unitlength,page=2]{controller_comparison4.pdf}}%
    \put(0.85376579,0.38567634){\color[rgb]{0,0,0}\makebox(0,0)[lt]{\lineheight{1.25}\smash{\begin{tabular}[t]{l}Detector\end{tabular}}}}%
    \put(0,0){\includegraphics[width=\unitlength,page=3]{controller_comparison4.pdf}}%
    \put(0.52863195,0.45312979){\color[rgb]{0,0,0}\rotatebox{-90.00000252}{\makebox(0,0)[lt]{\lineheight{1.25}\smash{\begin{tabular}[t]{l}F\end{tabular}}}}}%
    \put(0.52863195,0.43755384){\color[rgb]{0,0,0}\rotatebox{-90.00000252}{\makebox(0,0)[lt]{\lineheight{1.25}\smash{\begin{tabular}[t]{l}unction 2\end{tabular}}}}}%
    \put(0,0){\includegraphics[width=\unitlength,page=4]{controller_comparison4.pdf}}%
    \put(0.76660255,0.45521787){\color[rgb]{0,0,0}\rotatebox{-90.00000252}{\makebox(0,0)[lt]{\lineheight{1.25}\smash{\begin{tabular}[t]{l}Function N\end{tabular}}}}}%
    \put(0,0){\includegraphics[width=\unitlength,page=5]{controller_comparison4.pdf}}%
    \put(0.28463695,0.20164311){\color[rgb]{1,1,1}\makebox(0,0)[lt]{\lineheight{1.25}\smash{\begin{tabular}[t]{l}DNN\end{tabular}}}}%
    \put(0,0){\includegraphics[width=\unitlength,page=6]{controller_comparison4.pdf}}%
    \put(0.30077725,0.45369414){\color[rgb]{0,0,0}\rotatebox{-90.00000252}{\makebox(0,0)[lt]{\lineheight{1.25}\smash{\begin{tabular}[t]{l}F\end{tabular}}}}}%
    \put(0.30077725,0.43811819){\color[rgb]{0,0,0}\rotatebox{-90.00000252}{\makebox(0,0)[lt]{\lineheight{1.25}\smash{\begin{tabular}[t]{l}unction 1\end{tabular}}}}}%
    \put(0,0){\includegraphics[width=\unitlength,page=7]{controller_comparison4.pdf}}%
    \put(0.45285439,0.01413633){\color[rgb]{0,0,0}\makebox(0,0)[lt]{\lineheight{1.25}\smash{\begin{tabular}[t]{l}Controller\end{tabular}}}}%
    \put(0,0){\includegraphics[width=\unitlength,page=8]{controller_comparison4.pdf}}%
    \put(0.28223848,0.19748105){\color[rgb]{0,0,0}\makebox(0,0)[lt]{\lineheight{1.25}\smash{\begin{tabular}[t]{l}FCs\end{tabular}}}}%
    \put(0,0){\includegraphics[width=\unitlength,page=9]{controller_comparison4.pdf}}%
    \put(0.51053055,0.19767858){\color[rgb]{0,0,0}\makebox(0,0)[lt]{\lineheight{1.25}\smash{\begin{tabular}[t]{l}FCs\end{tabular}}}}%
    \put(0,0){\includegraphics[width=\unitlength,page=10]{controller_comparison4.pdf}}%
    \put(0.74806378,0.19588678){\color[rgb]{0,0,0}\makebox(0,0)[lt]{\lineheight{1.25}\smash{\begin{tabular}[t]{l}FCs\end{tabular}}}}%
    \put(0.1932129,0.12321312){\color[rgb]{0,0,0}\makebox(0,0)[lt]{\lineheight{1.25}\smash{\begin{tabular}[t]{l}Encoder\end{tabular}}}}%
    \put(0,0){\includegraphics[width=\unitlength,page=11]{controller_comparison4.pdf}}%
    \put(0.42153318,0.1213931){\color[rgb]{0,0,0}\makebox(0,0)[lt]{\lineheight{1.25}\smash{\begin{tabular}[t]{l}Encoder\end{tabular}}}}%
    \put(0,0){\includegraphics[width=\unitlength,page=12]{controller_comparison4.pdf}}%
    \put(0.66037852,0.12400321){\color[rgb]{0,0,0}\makebox(0,0)[lt]{\lineheight{1.25}\smash{\begin{tabular}[t]{l}Encoder\end{tabular}}}}%
    \put(0,0){\includegraphics[width=\unitlength,page=13]{controller_comparison4.pdf}}%
    \put(0.14871422,0.18880423){\color[rgb]{0,0,0}\rotatebox{-90.00000252}{\makebox(0,0)[lt]{\lineheight{1.25}\smash{\begin{tabular}[t]{l}F\end{tabular}}}}}%
    \put(0.14871422,0.17322828){\color[rgb]{0,0,0}\rotatebox{-90.00000252}{\makebox(0,0)[lt]{\lineheight{1.25}\smash{\begin{tabular}[t]{l}unction 1\end{tabular}}}}}%
    \put(0,0){\includegraphics[width=\unitlength,page=14]{controller_comparison4.pdf}}%
    \put(0.37896739,0.18616592){\color[rgb]{0,0,0}\rotatebox{-90.00000252}{\makebox(0,0)[lt]{\lineheight{1.25}\smash{\begin{tabular}[t]{l}F\end{tabular}}}}}%
    \put(0.37896739,0.17058997){\color[rgb]{0,0,0}\rotatebox{-90.00000252}{\makebox(0,0)[lt]{\lineheight{1.25}\smash{\begin{tabular}[t]{l}unction 2\end{tabular}}}}}%
    \put(0,0){\includegraphics[width=\unitlength,page=15]{controller_comparison4.pdf}}%
    \put(0.6192377,0.19022921){\color[rgb]{0,0,0}\rotatebox{-90.00000252}{\makebox(0,0)[lt]{\lineheight{1.25}\smash{\begin{tabular}[t]{l}Function N\end{tabular}}}}}%
    \put(0,0){\includegraphics[width=\unitlength,page=16]{controller_comparison4.pdf}}%
  \end{picture}%
\endgroup%

%% file: figs/controller_comparison.pdf_tex
\begingroup%
  \makeatletter%
  \providecommand\color[2][]{%
    \errmessage{(Inkscape) Color is used for the text in Inkscape, but the package 'color.sty' is not loaded}%
    \renewcommand\color[2][]{}%
  }%
  \providecommand\transparent[1]{%
    \errmessage{(Inkscape) Transparency is used (non-zero) for the text in Inkscape, but the package 'transparent.sty' is not loaded}%
    \renewcommand\transparent[1]{}%
  }%
  \providecommand\rotatebox[2]{#2}%
  \newcommand*\fsize{\dimexpr\f@size pt\relax}%
  \newcommand*\lineheight[1]{\fontsize{\fsize}{#1\fsize}\selectfont}%
  \ifx\svgwidth\undefined%
    \setlength{\unitlength}{712.0756073bp}%
    \ifx\svgscale\undefined%
      \relax%
    \else%
      \setlength{\unitlength}{\unitlength * \real{\svgscale}}%
    \fi%
  \else%
    \setlength{\unitlength}{\svgwidth}%
  \fi%
  \global\let\svgwidth\undefined%
  \global\let\svgscale\undefined%
  \makeatother%
  \begin{picture}(1,0.49143613)%
    \lineheight{1}%
    \setlength\tabcolsep{0pt}%
    \put(0,0){\includegraphics[width=\unitlength,page=1]{controller_comparison.pdf}}%
    \put(0.41890011,0.45897332){\color[rgb]{0,0,0}\makebox(0,0)[lt]{\lineheight{1.25}\smash{\begin{tabular}[t]{l}Differentiable ISP\end{tabular}}}}%
    \put(0,0){\includegraphics[width=\unitlength,page=2]{controller_comparison.pdf}}%
    \put(0.85322195,0.33881755){\color[rgb]{0,0,0}\makebox(0,0)[lt]{\lineheight{1.25}\smash{\begin{tabular}[t]{l}Detector\end{tabular}}}}%
    \put(0,0){\includegraphics[width=\unitlength,page=3]{controller_comparison.pdf}}%
    \put(0.87949725,0.14183004){\color[rgb]{0,0,0}\makebox(0,0)[lt]{\lineheight{1.25}\smash{\begin{tabular}[t]{l}SFB\end{tabular}}}}%
    \put(0,0){\includegraphics[width=\unitlength,page=4]{controller_comparison.pdf}}%
    \put(0.52879012,0.40609973){\color[rgb]{0,0,0}\rotatebox{-90.00000252}{\makebox(0,0)[lt]{\lineheight{1.25}\smash{\begin{tabular}[t]{l}F\end{tabular}}}}}%
    \put(0.52879012,0.39059576){\color[rgb]{0,0,0}\rotatebox{-90.00000252}{\makebox(0,0)[lt]{\lineheight{1.25}\smash{\begin{tabular}[t]{l}unction 2\end{tabular}}}}}%
    \put(0,0){\includegraphics[width=\unitlength,page=5]{controller_comparison.pdf}}%
    \put(0.76563294,0.4081922){\color[rgb]{0,0,0}\rotatebox{-90.00000252}{\makebox(0,0)[lt]{\lineheight{1.25}\smash{\begin{tabular}[t]{l}Function N\end{tabular}}}}}%
    \put(0,0){\includegraphics[width=\unitlength,page=6]{controller_comparison.pdf}}%
    \put(0.28588053,0.15578923){\color[rgb]{1,1,1}\makebox(0,0)[lt]{\lineheight{1.25}\smash{\begin{tabular}[t]{l}DNN\end{tabular}}}}%
    \put(0,0){\includegraphics[width=\unitlength,page=7]{controller_comparison.pdf}}%
    \put(0.51227502,0.15477811){\color[rgb]{1,1,1}\makebox(0,0)[lt]{\lineheight{1.25}\smash{\begin{tabular}[t]{l}DNN\end{tabular}}}}%
    \put(0,0){\includegraphics[width=\unitlength,page=8]{controller_comparison.pdf}}%
    \put(0.74928636,0.15571902){\color[rgb]{1,1,1}\makebox(0,0)[lt]{\lineheight{1.25}\smash{\begin{tabular}[t]{l}DNN\end{tabular}}}}%
    \put(0,0){\includegraphics[width=\unitlength,page=9]{controller_comparison.pdf}}%
    \put(0.3735538,0.15578923){\color[rgb]{1,1,1}\makebox(0,0)[lt]{\lineheight{1.25}\smash{\begin{tabular}[t]{l}DNN\end{tabular}}}}%
    \put(0,0){\includegraphics[width=\unitlength,page=10]{controller_comparison.pdf}}%
    \put(0.59841755,0.15524154){\color[rgb]{1,1,1}\makebox(0,0)[lt]{\lineheight{1.25}\smash{\begin{tabular}[t]{l}DNN\end{tabular}}}}%
    \put(0,0){\includegraphics[width=\unitlength,page=11]{controller_comparison.pdf}}%
    \put(0.30196028,0.40671764){\color[rgb]{0,0,0}\rotatebox{-90.00000252}{\makebox(0,0)[lt]{\lineheight{1.25}\smash{\begin{tabular}[t]{l}F\end{tabular}}}}}%
    \put(0.30196028,0.39117154){\color[rgb]{0,0,0}\rotatebox{-90.00000252}{\makebox(0,0)[lt]{\lineheight{1.25}\smash{\begin{tabular}[t]{l}unction 1\end{tabular}}}}}%
    \put(0,0){\includegraphics[width=\unitlength,page=12]{controller_comparison.pdf}}%
    \put(0.53807284,0.01448402){\color[rgb]{0,0,0}\makebox(0,0)[lt]{\lineheight{1.25}\smash{\begin{tabular}[t]{l}Controller\end{tabular}}}}%
    \put(0,0){\includegraphics[width=\unitlength,page=13]{controller_comparison.pdf}}%
    \put(0.28350718,0.15373889){\color[rgb]{0,0,0}\makebox(0,0)[lt]{\lineheight{1.25}\smash{\begin{tabular}[t]{l}FCs\end{tabular}}}}%
    \put(0,0){\includegraphics[width=\unitlength,page=14]{controller_comparison.pdf}}%
    \put(0.50990168,0.15272776){\color[rgb]{0,0,0}\makebox(0,0)[lt]{\lineheight{1.25}\smash{\begin{tabular}[t]{l}FCs\end{tabular}}}}%
    \put(0,0){\includegraphics[width=\unitlength,page=15]{controller_comparison.pdf}}%
    \put(0.74691302,0.1537108){\color[rgb]{0,0,0}\makebox(0,0)[lt]{\lineheight{1.25}\smash{\begin{tabular}[t]{l}FCs\end{tabular}}}}%
    \put(0,0){\includegraphics[width=\unitlength,page=16]{controller_comparison.pdf}}%
    \put(0.37113832,0.15373889){\color[rgb]{0,0,0}\makebox(0,0)[lt]{\lineheight{1.25}\smash{\begin{tabular}[t]{l}FCs\end{tabular}}}}%
    \put(0,0){\includegraphics[width=\unitlength,page=17]{controller_comparison.pdf}}%
    \put(0.59604421,0.15320524){\color[rgb]{0,0,0}\makebox(0,0)[lt]{\lineheight{1.25}\smash{\begin{tabular}[t]{l}FCs\end{tabular}}}}%
    \put(0,0){\includegraphics[width=\unitlength,page=18]{controller_comparison.pdf}}%
    \put(0.87956747,0.06945009){\color[rgb]{0,0,0}\makebox(0,0)[lt]{\lineheight{1.25}\smash{\begin{tabular}[t]{l}FCs\end{tabular}}}}%
    \put(0,0){\includegraphics[width=\unitlength,page=19]{controller_comparison.pdf}}%
  \end{picture}%
\endgroup%